\documentclass[10pt,twocolumn,letterpaper]{article}

\usepackage{cvpr}

\usepackage{booktabs}
\usepackage{colortbl}
\usepackage{pifont}
\usepackage{multirow}
\usepackage{array}
\usepackage{tabularx}
\newcommand{\cmark}{\ding{51}}
\newcommand{\xmark}{\ding{55}}
\newcommand{\ptmark}{$\triangle$}

\definecolor{qcSand}{RGB}{252,246,232}
\definecolor{qcSandBest}{RGB}{244,232,204}
\definecolor{qcTeal}{RGB}{226,239,236}
\definecolor{qcTealBest}{RGB}{198,224,216}
\definecolor{qcSlate}{RGB}{232,237,242}
\definecolor{qcSlateBest}{RGB}{210,221,232}
\definecolor{qcGold}{RGB}{243,236,214}
\definecolor{qcClay}{RGB}{246,236,230}
\usepackage{amsmath,amssymb}
\usepackage{graphicx}
\DeclareGraphicsExtensions{.pdf}
\usepackage{tikz}
\usetikzlibrary{arrows.meta,positioning,fit}
\usepackage{microtype}

\makeatletter
\def\cvprsection{\@startsection {section}{1}{\z@}
   {-8pt plus -1pt minus -1pt}{5pt} {\large\bf}}
\def\cvprsubsection{\@startsection {subsection}{2}{\z@}
   {-6pt plus -1pt minus -1pt}{3pt} {\elvbf}}
\makeatother
\renewcommand{\paragraph}[1]{\vspace{0.2em}\noindent{\tenbf #1}\hspace{0.35em}}
\definecolor{cvprblue}{rgb}{0.21,0.49,0.74}
\usepackage[pagebackref,breaklinks,colorlinks,allcolors=cvprblue]{hyperref}

\def\confName{CVPR}
\def\confYear{2027}

\makeatletter
\def\@maketitle{%
  \newpage
  \null
  \vskip 0.12in
  \begin{center}%
    {\LARGE\bfseries \@title \par}%
    \vskip 16pt
    {\large \@author \par}%
  \end{center}%
  \vskip 6pt
}
\makeatother

\title{PhoneValue-Bench: A Dual-Ground-Truth Scorecard Benchmark for Smartphone Recommerce Inspection}

\author{%
Zihan Deng$^{1}$ \quad Chuanzhi Xu$^{2}$ \quad Junjie Huang$^{3}$ \quad Xiaozhen Zhong$^{4}$ \quad Du Tiantian$^{1}$\\[0.7em]
{\normalsize $^{1}$The University of Hong Kong \quad $^{2}$The University of Sydney}\\[0.28em]
{\normalsize $^{3}$University of Science and Technology of China}\\[0.28em]
{\normalsize $^{4}$University of Electronic Science and Technology of China}%
}

\begin{document}
\maketitle
\begin{abstract}
Remote smartphone recommerce is scaling quickly, and locking a repurchase offer from seller photos and diagnostic screens before the handset arrives has become a central operational task.
Existing multimodal resources mainly target defect localization, catalog attributes, or screen reading.
They overlook payout-facing scorecards with dual cosmetic and operational ground truth under a declared input contract.
This gap stems from two challenges: (1)~the lack of a large weakly labeled trade-in corpus kept separate from a compact audited ranking cut, and (2)~architectures that either flatten unordered photo bags into one prompt or collapse evaluation into a single price scalar.
To address these challenges, we propose PhoneValue-Bench, a dual-ground-truth recommerce framework with PhoneValue-Corpus for weak training on 21{,}550 orders and 278{,}748 images, and PV-Rank as the only leaderboard on 335 quality-gated orders.
Systems are ranked by CardScore under LeadGate, with Cosmetic Fill and Operational Fill as diagnostics, while grade and price stay off the board.
Notably, ValueAgent is a six-stage perception--fusion reference that canonicalizes views, detects regional defects, reads screenshots, and fuses evidence under warehouse-style rules, so bag-of-images prompting is not the only evaluation unit.
On the same PV-Rank cut and frozen scorer, this full-system and input-contract comparison finds ValueAgent at 84.1 CardScore against the strongest Flat Vision-Language Model at 69.9, with the largest reported gap in CosFill, while OpsFill remains limited under the two declared contracts.
CardScore measures field-fill completeness, not quote accuracy.
\end{abstract}

\section{Introduction}
\label{sec:intro}

\begin{figure}[t]
\centering
\includegraphics[width=\linewidth]{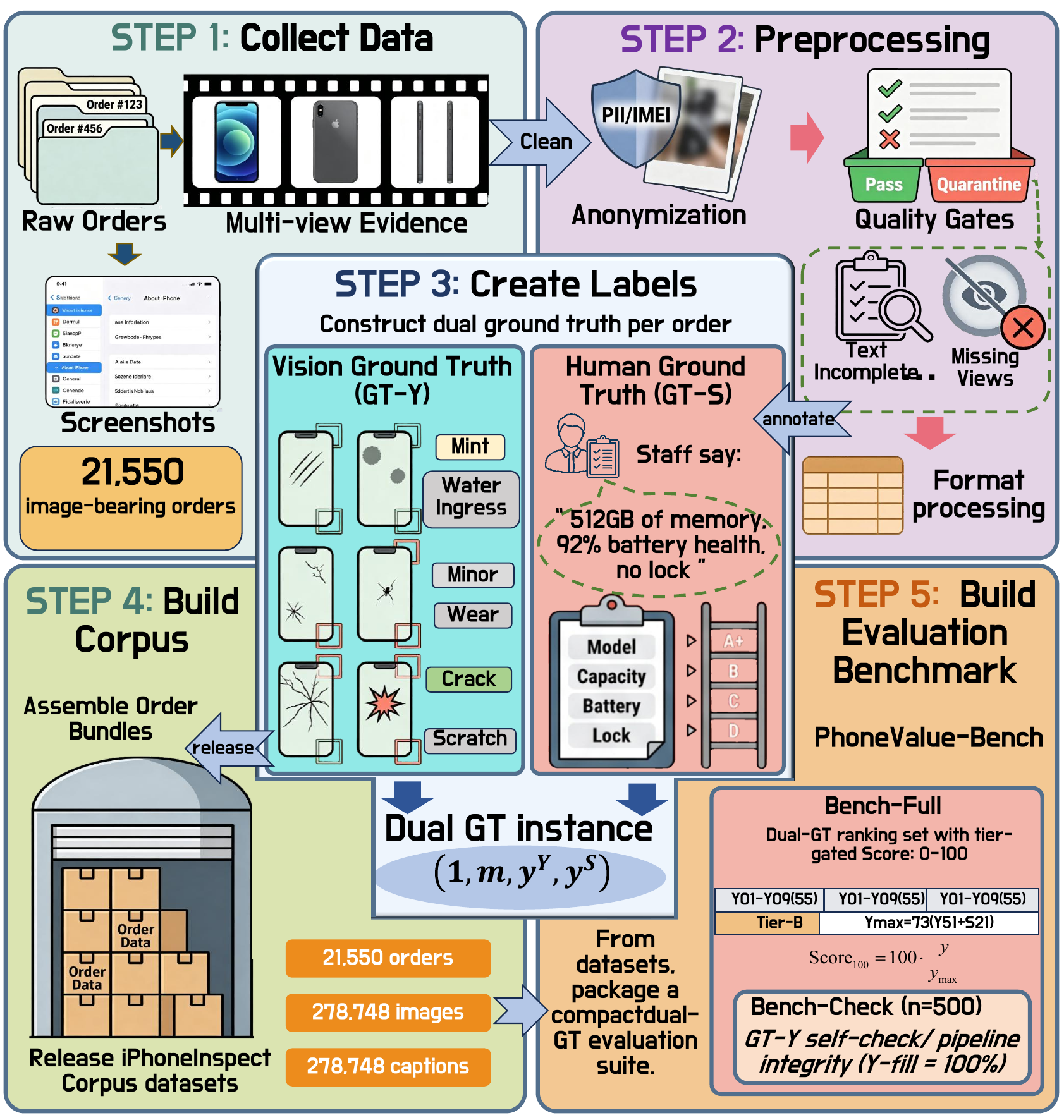}
\caption{PhoneValue construction pipeline from the final production release.
Step~1 collects 21{,}550 image-bearing orders with multi-view photos and screenshots.
Step~2 anonymizes identifiers and applies Pass or Quarantine gates.
Step~3 builds dual Cosmetic Ground Truth boxes and Operational Ground Truth staff fields.
Step~4 releases PhoneValue-Corpus with 278{,}748 weakly captioned images.
Step~5 packages PV-Audit with 500 orders for integrity checks and PV-Rank with 335 orders as the only CardScore leaderboard.}
\label{fig:pipeline}
\end{figure}

Remote smartphone recommerce is growing rapidly.
Before a handset arrives, buyers must lock a repurchase band from seller photographs and a diagnostic screen.
Warehouse staff answer that request by filling a payout scorecard: which exterior views exist, where defects sit, which model and capacity the screen reports, whether locks or water history block the quote, and which cosmetic ceiling applies.
A missed side view or a wrong blocker can move the offer by a full grade tier even when a crack box looks plausible.
Assessing multimodal systems for this desk therefore needs more than a fluent caption or a single price scalar.
It needs a scorecard that a warehouse can audit under a known input contract.

Existing public resources cover only pieces of that desk.
Industrial and logistics resources such as MVTec AD, VisA, Kaputt, and the Smartphone Screen Glass Dataset evaluate visual defects in their respective capture settings~\cite{mvtec2019,visa2022,kaputt2025,ssgd2023}.
E-commerce attribute suites such as M5Product, SMARTAVE, and ImplicitAVE recover catalog values from product media~\cite{dong2022m5product,wang2022smartave,zou2024implicitave}.
Document and screen benchmarks such as DocVQA, ScreenQA, MMBench, and MMMU test reading or broad multimodal competence~\cite{docvqa2021,screenqa2025,mmbench2024,mmmu2024}.
These lines are useful, but they stop short of the full payout card.
A correct box on one pane still leaves storage, lock, and channel unread.
A correct listing attribute still leaves regional wear unscored.
None of these resources release dual cosmetic and operational ground truth, isolate a large weak corpus from ranking gold, and rank systems with an executable scorecard that withholds price from the leaderboard.

The deeper problem is the evaluation unit itself.
Unordered seller bags do not tell a model which side a scratch belongs to.
View canonicalization and detection may help recover view coverage and regional defect locations.
Operational fields live in screens and staff vocabulary, so a bag-of-images prompt can read capacity yet still miss blockers, channel, and exact grade strings.
Price regression then collapses both components into one number and hides which field failed.
If training weak labels mix into the ranking cut, a staged detector can also look strong for the wrong reason.
Public progress is therefore blocked by two coupled challenges: (1)~the lack of a dual-ground-truth recommerce resource that separates a large weakly labeled corpus from a compact audited ranking cut, and (2)~the need for two evaluation contracts under one scorer, while making their different inputs explicit.

In this paper, we present PhoneValue-Bench as that resource, as shown in Figure~\ref{fig:pipeline}.
PhoneValue-Corpus supplies 21{,}550 orders and 278{,}748 weakly captioned images for pretraining.
PV-Rank supplies 335 quality-gated orders with Cosmetic Ground Truth on the views and Operational Ground Truth on staff fields as the only public leaderboard.
CardScore under LeadGate, the fixed ranking gate with 51 cosmetic and 22 operational points, averages gated item fill on that cut.
Cosmetic Fill and Operational Fill serve as diagnostics, while dashboard grade and repurchase price stay off the board.
We evaluate two declared contracts on the same PV-Rank orders and scorer, but they have different input privileges.
Flat Vision-Language Model, written FlatVLM, is restricted to a single-turn image bag.
ValueAgent is a six-stage perception--fusion reference with view canonicalization, detection, screenshot reading, and deterministic fusion.
ValueAgent can use detector crops, repair history, and Expert Knowledge Base resources unavailable to FlatVLM; their score difference is therefore not a controlled comparison of model architectures.
On PV-Rank, this full-system and input-contract comparison finds ValueAgent at 84.1 CardScore with CosFill 97.0 and OpsFill 54.1, against the strongest FlatVLM at 69.9.
The largest component of the score gap is cosmetic under this setup. Grade strings, channel, and blockers remain weak.
Our contributions are as follows.

\textbf{PhoneValue-Bench.}
We construct a dual-ground-truth recommerce scorecard on 335 quality-gated PV-Rank orders under LeadGate: Cosmetic Ground Truth on the views, Operational Ground Truth on staff fields, and CardScore as a gated 0 to 100 leaderboard with CosFill and OpsFill diagnostics, while a large weakly labeled Corpus stays held out from ranking gold.

\textbf{ValueAgent.}
We propose ValueAgent, a six-stage perception--fusion reference that canonicalizes views, detects regional defects, reads screenshots, and fuses warehouse-style evidence under a declared contract, with FlatVLM as the single-turn image-bag baseline under the same scorer.

\textbf{Public protocol and findings.}
Under that full-system and input-contract comparison, ValueAgent reaches 84.1 CardScore against the strongest FlatVLM at 69.9. The largest reported difference is in CosFill. Operational fields such as grade strings, channel, and blockers remain weak on this compact Rank cut.

\section{Related Work}

\begin{table*}[t]
\centering\scriptsize
\setlength{\tabcolsep}{1.8pt}
\renewcommand{\arraystretch}{0.96}
\caption{Closest public resources versus PhoneValue-Bench.
A checkmark means coverage usable for ranking. A triangle means partial coverage only.}
\label{tab:positioning}
\begin{tabularx}{\textwidth}{@{}X*{7}{c}@{}}
\toprule
\textbf{Dimension} & \textbf{DocVQA} & \textbf{ScreenQA} & \textbf{MMBench} & \textbf{SSGD} & \textbf{VisA} & \textbf{M5Product} & \textbf{Ours} \\
\midrule
\rowcolor{qcSand}
Large training or crawl set & \ptmark & \ptmark & \xmark & \ptmark & \ptmark & \cmark & \ptmark \\
\rowcolor{qcTealBest}
Compact audited dual ground truth & \xmark & \xmark & \xmark & \xmark & \xmark & \xmark & \cmark \\
\rowcolor{qcSand}
Multi-view device cosmetics & \xmark & \xmark & \xmark & \ptmark & \ptmark & \ptmark & \cmark \\
\rowcolor{qcSand}
Diagnostic screenshot fields & \xmark & \xmark & \xmark & \xmark & \xmark & \xmark & \cmark \\
\rowcolor{qcTealBest}
Operational risk and blocker fields & \xmark & \xmark & \xmark & \xmark & \xmark & \xmark & \ptmark \\
\rowcolor{qcTealBest}
Gated payout scorecard & \xmark & \xmark & \xmark & \xmark & \xmark & \xmark & \cmark \\
\rowcolor{qcTealBest}
Separate weak recommerce corpus & \xmark & \xmark & \xmark & \xmark & \xmark & \xmark & \cmark \\
\rowcolor{qcTealBest}
Staged agent reference protocol & \xmark & \xmark & \xmark & \xmark & \xmark & \xmark & \cmark \\
\midrule
\rowcolor{qcSlate}
Primary target & doc VQA & screen QA & VLM suite & factory glass & industrial AD & e-com.\ pretrain & recommerce card \\
\bottomrule
\end{tabularx}
\end{table*}

\subsection{Industrial defects and e-commerce attributes}
Recommerce scoring sits between factory inspection and catalog understanding.
Early industrial anomaly datasets such as MVTec AD and VisA annotate controlled workstation imagery with pixel-level defects~\cite{mvtec2019,visa2022}.
Later logistics and phone-glass resources such as Kaputt and the Smartphone Screen Glass Dataset extend localization to more realistic capture settings, but their targets differ from a complete operational inspection form~\cite{kaputt2025,ssgd2023}.
Deep detectors trained on those corpora improve box quality.
They still omit unordered seller multi-views, diagnostic screens, and payout-facing fields such as activation locks or channel flags.
On the catalog side, M5Product scales multimodal e-commerce pretraining~\cite{dong2022m5product}, while SMARTAVE and ImplicitAVE extract structured or implied attribute values from product images~\cite{wang2022smartave,zou2024implicitave}.
These resources study product representations and attributes using catalog text and visual evidence.
However, these methods rely on controlled defect panes or catalog attributes, whereas remote recommerce requires multi-view cosmetic consistency and operational risk fields under one warehouse scorecard.
PhoneValue therefore treats localization and attribute extraction as related skills, not as substitutes for recommerce ranking.
Table~\ref{tab:positioning} summarizes the gap against common neighbors.

\subsection{Screen reading, mobile UI, and agents}
A second neighboring line studies screens and agents, not factory panes.
Document and screen reading evolved from DocVQA and TextVQA to ScreenQA and broad multimodal suites such as MMBench and MMMU~\cite{docvqa2021,textvqa2019,screenqa2025,mmbench2024,mmmu2024}.
Mobile and web UI corpora and agents such as Rico, Android in the Wild, AppAgent, SeeAct, and GUI Odyssey study interface design or device control~\cite{rico2017,aitw2023,appagent2025,seeact2024,guiodyssey2025}.
Screenshot reading supports model, capacity, and battery extraction in ValueAgent Stages~4--5. Prior agent work provides related methods for multi-step UI interaction.
Existing studies mainly focus on reading fidelity, open-domain competence, or action success, while paying limited attention to dual Cosmetic and Operational Ground Truth, Corpus versus Rank isolation, and a frozen payout scorecard that holds price off the leaderboard.
Proprietary warehouse stacks often fuse vision and forms in production, but they rarely publish those three elements together.
PhoneValue-Bench is built so that Flat Vision-Language Model and ValueAgent can be compared on that missing unit: one order, one scorecard, one declared contract.

\section{The PhoneValue Benchmark}
\label{sec:protocol}

\begin{figure*}[t]
\centering
\includegraphics[width=\textwidth]{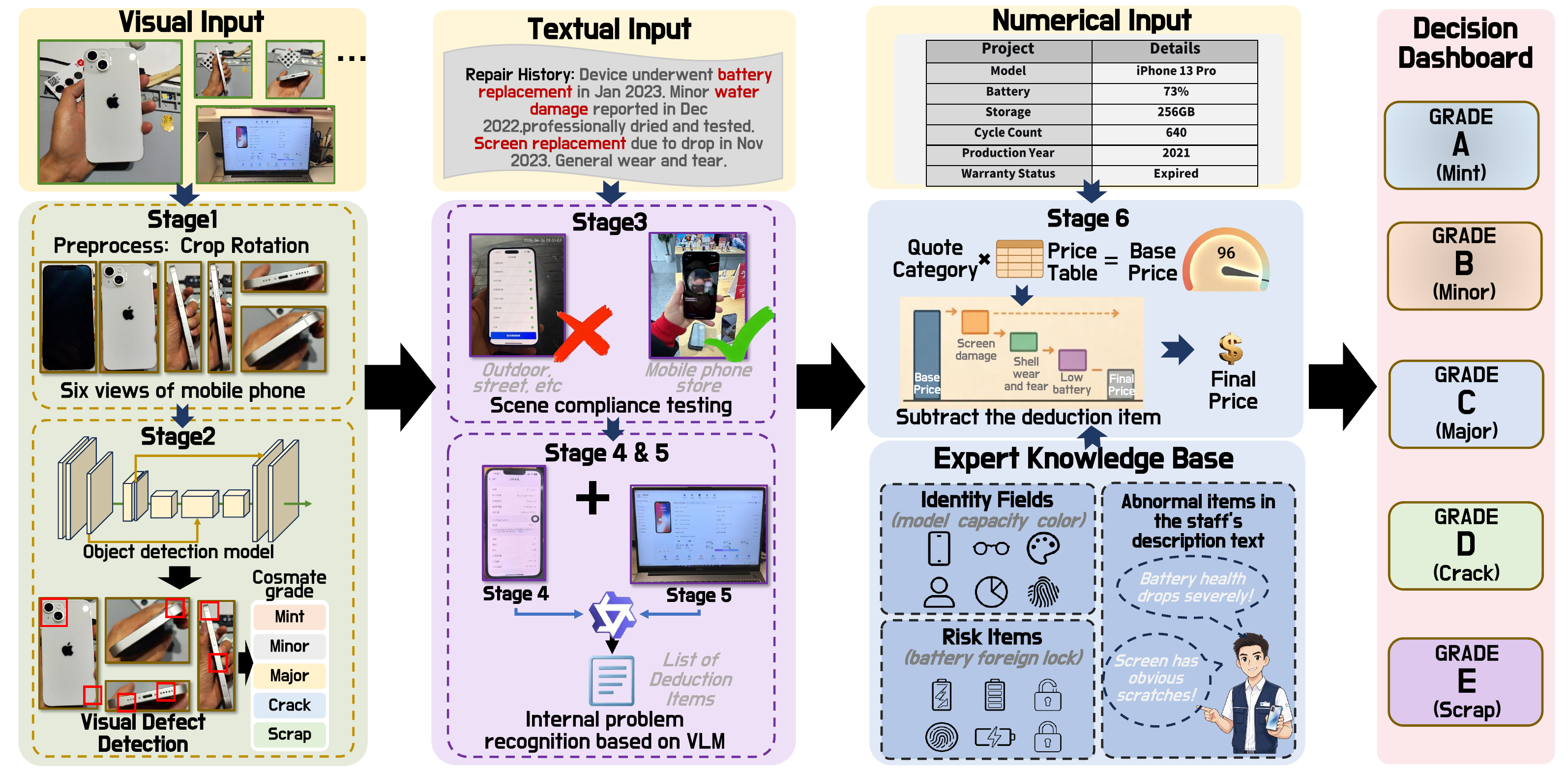}
\caption{ValueAgent reference pipeline from the final production release.
Visual, repair-history, and structured device fields enter Stages~1--2 for six-view crops and defect detection, Stage~3 for scene compliance, and Stages~4--5 for About and diagnostic screenshot reading.
An Expert Knowledge Base supports fusion, and Stage~6 computes off-leaderboard grade and price.
FlatVLM sees only the raw image bag under a single-turn schema.
Both contracts emit the same scorecard for CardScore. Grade and price stay off the public rank.}
\label{fig:method}
\end{figure*}

We first fix the public evaluation unit, then define CardScore, then state the two system contracts, and finally describe how Corpus and Rank are isolated.
The public unit in PhoneValue is a payout scorecard tied to one intake order.
Dual ground truth, as used in the title and abstract, means two complementary references on that same order, not two independent copies of one label set.
CardScore ranks how completely the card is filled, not how accurately a repurchase quote is priced.
Repurchase price is computed later and never enters the leaderboard.
Submitters must declare whether they run FlatVLM on the raw image bag or ValueAgent with staged perception, because detector crops and Expert Knowledge Base dumps are privileged relative to FlatVLM.

\paragraph{Orders and dual ground truth.}
A PV-Rank item is a quality-gated order with seller upload $X$ and dual reference $a^\star=(c^\star,o^\star)$ on the same intake record.
Cosmetic Ground Truth $c^\star$, CosGT, records view presence, regional defect boxes, and cosmetic severity.
Operational Ground Truth $o^\star$, OpsGT, records model, capacity, masked International Mobile Equipment Identity last-4, grade strings, risk flags, and no-quote blockers, filled from diagnostic screenshots and intake forms on that same order.
The two references can disagree. They are not interchangeable gold copies.
When they disagree, O-items follow staff priority for O04--O10 and C-items keep vision boxes for C04--C08.
The system returns scorecard $\hat{y}$ in the same schema.
Orders that fail Step~2 for missing mandatory views, incomplete staff text, or unreadable screenshots never receive dual ground truth and never enter PV-Rank.
Their robustness behavior is discussed only in the appendix.

\paragraph{CardScore.}
Active items cover view coverage, regional defect matching on display, body and frame, ports, and camera sides, grade agreement after a fixed taxonomy crosswalk, and operational fields for model, storage, battery-related flags, lock and biometrics, channel, and blockers.
Table~\ref{tab:item-budget} lists LeadGate maxima for the public board.
If an item has no evidence on that order, its gate $g_j$ is off and the item leaves both the numerator and the denominator of the ratio.
With awarded points $\hat{s}_j\in[0,M_j]$,
\begin{equation}
\mathrm{CardScore}=100\cdot\frac{\sum_j g_j\hat{s}_j}{\sum_j g_j M_j}.
\label{eq:score}
\end{equation}
On PV-Rank under LeadGate, global gates deactivate C09, O11--O12, and K01--K02 for every order, giving nominal maxima of 51 cosmetic and 22 operational points. When all these items are active,
\begin{equation}
\mathrm{CardScore}=\frac{51\cdot\mathrm{CosFill}+22\cdot\mathrm{OpsFill}}{73},
\label{eq:leadgate}
\end{equation}
where Cosmetic Fill and Operational Fill in Eq.~\eqref{eq:fills} reuse the same gates and maxima as Eq.~\eqref{eq:score}:
\begin{equation}
\mathrm{CosFill}=100\frac{\sum_{j\in\mathcal{C}} g_j\hat{s}_j}{\sum_{j\in\mathcal{C}} g_j M_j},\qquad
\mathrm{OpsFill}=100\frac{\sum_{j\in\mathcal{O}} g_j\hat{s}_j}{\sum_{j\in\mathcal{O}} g_j M_j}.
\label{eq:fills}
\end{equation}
Both fills are percentages. Eq.~\eqref{eq:leadgate} holds only when the active maxima equal 51 and 22; otherwise Eq.~\eqref{eq:score} uses order-specific denominators.
Reported CardScore, CosFill, and OpsFill are order-level means of these per-order gated ratios, not a single field-level micro pool that first sums points across all orders and then divides once.
Per-order gates still deactivate O09 without channel metadata and may partially deactivate C04--C07 when a mandatory side view failed preprocessing.
Gates must be set from reference evidence, independently of model predictions. Inactive items contribute neither points nor maximum points.
Dashboard grade $g\in\{A,\ldots,E\}$ and repurchase price $p$ are computed by the same warehouse rules but remain off the leaderboard.
In short, CardScore measures scorecard fill, and price-band agreement is reported only as a secondary check on a stated subset and protocol.
Box matching for C04--C07 pairs predictions and CosGT greedily by class agreement and intersection over union at least $1/2$, then enters cosmetic points only.
The matcher is public and frozen with the scorer; it is not a second rank and does not use Rank boxes as detector training targets.

\begin{table}[t]
\centering\scriptsize
\setlength{\tabcolsep}{3pt}
\renewcommand{\arraystretch}{0.96}
\caption{LeadGate point budget on PV-Rank.
C09, O11--O12, and K01--K02 are globally inactive.}
\label{tab:item-budget}
\begin{tabularx}{\columnwidth}{@{}Xlc@{}}
\toprule
\textbf{Group} & \textbf{Items} & \textbf{Max} \\
\midrule
\rowcolor{qcSand}
Cosmetic views and counts & C01--C03 & 14 \\
\rowcolor{qcSand}
Regional defect matching & C04--C07 & 31 \\
\rowcolor{qcSand}
Cosmetic grade agreement & C08 & 6 \\
\rowcolor{qcTeal}
Operational identity & O01--O03 & 6 \\
\rowcolor{qcTeal}
Grade string and ceiling & O04--O05 & 9 \\
\rowcolor{qcTeal}
Risk flags and blockers & O06--O10 & 7 \\
\midrule
\rowcolor{qcSlateBest}
Active total & -- & 73 \\
\bottomrule
\end{tabularx}
\end{table}

\begin{table}[t]
\centering\scriptsize
\setlength{\tabcolsep}{3pt}
\renewcommand{\arraystretch}{0.96}
\caption{PV-Rank composition under LeadGate.
Severity and family strata are fixed before ranking.}
\label{tab:rank-mix}
\begin{tabularx}{\columnwidth}{@{}Xcc@{}}
\toprule
\textbf{Slice} & \textbf{$n$} & \textbf{Share} \\
\midrule
\rowcolor{qcSand}
Mint or excellent & 48 & 14.3\% \\
\rowcolor{qcSand}
Minor wear & 142 & 42.4\% \\
\rowcolor{qcClay}
Heavy or major & 97 & 29.0\% \\
\rowcolor{qcClay}
Crack or scrap & 48 & 14.3\% \\
\midrule
\rowcolor{qcTeal}
Family iPhone 11--13 & 156 & 46.6\% \\
\rowcolor{qcTeal}
Family iPhone 14--15 & 89 & 26.6\% \\
\rowcolor{qcTeal}
Family other or Android & 90 & 26.9\% \\
\midrule
\rowcolor{qcSlateBest}
PV-Rank total & 335 & 100\% \\
\bottomrule
\end{tabularx}
\end{table}

\paragraph{Declared input contracts.}
With the scorer fixed, we next declare what each system may see.
Figure~\ref{fig:method} contrasts the two ranking contracts.
Both contracts start from the same post-Step~2 image bag, including appearance photographs and diagnostic screenshots, with no fixed three-image cap and no different raw-file selection for the two systems.
Missing photographs stay missing, excess images stay in the bag, and appearance versus diagnostic priority is used only for later routing, as specified in the appendix.
FlatVLM then sees that shared bag under a single-turn English schema at temperature~0, without detector crops, tools, Expert Knowledge Base fusion, or the price table.
ValueAgent may run view canonicalization, an external detector, screenshot vision-language model stages, repair-history text through the Expert Knowledge Base, and deterministic fusion.
The later comparison is a full-system and input-contract comparison on the same files.
Both still submit only $\hat{y}$ to CardScore.
Reading OpsGT at test time is forbidden on both boards.
Heuristic optical character recognition and single-image vision-language model runs are modality probes under the same scorer.
PV-Audit oracle replay checks CosGT completeness and is not a FlatVLM peer.

\paragraph{Corpus, Rank, and isolation.}
Figure~\ref{fig:pipeline} and the construction steps turn operational orders into Corpus and Rank with hard boundaries.
Step~1 collects 21{,}550 image-bearing order folders.
Step~2 anonymizes personal identifiers and International Mobile Equipment Identity strings, then applies Pass and Quarantine gates so incomplete staff text or missing mandatory views never receive dual ground truth for ranking.
Step~3 builds dual ground truth with eight warehouse-trained annotators under a written guideline.
A 10\% double-label slice is labeled independently and then adjudicated by a senior inspector, as detailed in the appendix together with class definitions and remaining agreement gaps.
On that audit subset, Cohen's $\kappa$ is 0.81 for CosGT severity, 0.88 for view presence, 0.89 for model, 0.94 for capacity, and 0.86 for lock.
About 7.4\% of dual-labeled items need adjudication.
Step~4 releases PhoneValue-Corpus with weak English captions and detector sidecars for training only.
Each Corpus image carries one of four English caption styles aimed at narrative notes, technical detector text, concise retrieval, and audit mismatches, with mean length near 85 words.
Step~5 packages PV-Audit with 500 orders for CosGT integrity under oracle replay and PV-Rank with 335 orders as the only public ranking set.
An audit on order identifiers and on 256-bit secure hashes of the images finds no overlap between the reported training runs and the 335-order cut.
That audit does not exclude near-duplicate images, crops, compressed or re-shot frames, or shared product models, which exact hashes cannot rule out.
Corpus weak labels are never ranking gold.
PV-Rank is stratified by cosmetic severity and device family as listed in Table~\ref{tab:rank-mix}, to retain variation in both cosmetic condition and device family.
Diagnostic-screenshot coverage on Rank exceeds 94\% and side-view coverage exceeds 96\%, which keeps most C04--C07 and screenshot-driven O-items active under LeadGate while still leaving a small residual of incomplete uploads for partial C01 credit.
Leaderboard results use this cut and the frozen scorer; Corpus statistics and PV-Audit results are reported separately.

\section{ValueAgent Reference Pipeline}
\label{sec:valueagent}

Having fixed the scorecard, we publish ValueAgent as the reference pipeline that fills it.
It maps intake $(X,m)$ to $\hat{y}$, and logs grade $g$ with price $p$ only for secondary audits.
Stages are swappable so detector, vision-language model backend, and price table can change without rewriting CardScore.
Figure~\ref{fig:method} and Table~\ref{tab:agent-alg} list the six stages.
Table~\ref{tab:contracts} contrasts what each ranking contract may consume.
Fusion rules and prompts sit in the appendix.

\paragraph{Problem formulation.}
A system under a declared contract emits $\hat{y}$ in the public schema.
CardScore is discrete in those fields, so we do not backpropagate through it.
ValueAgent is built so CosFill should fall when Stages~1--2 are removed, and OpsFill should fall when Stages~4--5 or fusion are removed, while FlatVLM remains the image-bag control on the same scorer.
Formally, ValueAgent factors the map as
\begin{equation}
\begin{aligned}
d&=\mathrm{Fuse}(B,\hat{y}_{\mathrm{grade}},s_{\mathrm{screen}},s_{\mathrm{diag}},\mathrm{KB},m),\\
\hat{y}&=\mathrm{MapToScorecard}(d),
\end{aligned}
\label{eq:pipeline}
\end{equation}
where $B$ and $\hat{y}_{\mathrm{grade}}$ come from Stages~1--2, $s_{\mathrm{screen}}$ and $s_{\mathrm{diag}}$ from Stages~4--5, and $\mathrm{Fuse}$ applies the deterministic priority sheet in Section~\ref{sec:fusion}.
FlatVLM skips $B$, $\mathrm{KB}$, and $\mathrm{Fuse}$, conditioning only on $X$ and a fixed schema prompt.
Eq.~\eqref{eq:pipeline} defines the reference pipeline. Each listed ablation changes one named factor on the same PV-Rank orders, with the scorer and backend held fixed. The resulting differences are associated with that configuration. They do not isolate architecture from input privilege, and they are not a same-privilege comparison against FlatVLM.

\paragraph{Stages 1--3: views, defects, and scene gate.}
Seller uploads arrive unordered and often poorly framed.
Stage~1 detects the phone region, then crops and rotates each image into six canonical views $V$: front, back, and four edges.
Canonicalization makes C01 for view coverage and regional scoring on C04--C07 well-defined.
Side scratches are scored on side views and are not confused with background clutter on a single front photo.
Without Stage~1, a bag-of-images system must identify views and predict coordinates directly from raw uploads. That contrast mixes view processing and detection with a change in the artifacts available to the system.
Missing views reduce C01 and may gate the matching regional item, so later stages do not invent wear on an unseen edge.
Stage~2 runs a pluggable detector, You Only Look Once version~8 in the reported runs, on $V$ and outputs boxes $B$ plus a detector cosmetic grade in $\{\mathrm{mint},\mathrm{minor},\mathrm{major},\mathrm{crack},\mathrm{scrap}\}$~\cite{yolov8}.
The detector is trained on PhoneValue-Corpus appearance exports with PV-Rank order IDs and image hashes held out.
It is not fine-tuned on CosGT boxes from the ranking cut.
Those buckets map to scorecard slots after the fixed taxonomy crosswalk used by C08.
They differ from the CosGT severity strings, so C08 credits mapped agreement, not raw string equality.
Stage~3 then tests scene compliance.
Outdoor or street backgrounds are flagged, and store-style captures pass.
The gate does not rewrite defects.
It suppresses clutter-induced false positives before later reasoning and acts as a precision lever, not the primary recall driver.
On PV-Rank the Stage~3 ablation costs only a few CardScore points, which matches that reading: cosmetics are already largely decided by Stages~1--2.

\paragraph{Stages 4--5: screenshot reading.}
After cosmetics are grounded in views and boxes, Stages~4--5 read the screen.
Stage~4 reads About and settings screenshots to extract identity fields such as model, capacity, and lock status.
Stage~5 reads diagnostic screenshots for battery health and related internal status.
Both stages output structured JSON that fills the corresponding operational scorecard items, merging multiple screenshots by field priority.
Repair-history text is not handled as a separate vision stage.
It is fed into the Expert Knowledge Base so that prior water events or replacements can override overly optimistic visual grades.
The production vision-language model backend used in Stages~4--5 is swappable.
Reported cosmetic gains primarily reflect Stages~1--2, not backend choice alone, which is why the FlatVLM contract freezes a single-turn image bag without those stages.
When a crop is obscured by glare, truncated, or unrelated to the UI, the stage leaves the field blank and does not invent digits, matching FlatVLM's unknown policy.

\paragraph{Fusion, scorecard mapping, and off-leaderboard pricing.}
\label{sec:fusion}
The Expert Knowledge Base stores identity priors, risk items for battery, foreign parts, and locks, and abnormal phrases mined from staff text.
Fusion applies deterministic priority rules over detector outputs, vision-language model JSON, and metadata.
Screenshot identity overrides conflicting view labels.
Cosmetic ceilings follow worst-region order: scrap, crack, major, minor, then mint.
Water or activation-lock evidence from any modality propagates to O08 and O10 blockers.
The scorer deactivates O09 when OpsGT omits channel; inference must not inspect OpsGT to choose a channel.
Risk cues worsen ceilings one-way: a clean photo cannot erase water or lock evidence from screen or repair text.
\texttt{MapToScorecard} then writes $\hat{y}$ without consulting proprietary numeric price lists, keeping leaderboard scoring auditable.
If Stage~2 reports mint but Stage~5 reads water damage, fusion lowers the ceiling and may trigger blockers before C08 and O04--O05 are filled.
Stage~6 computes price and dashboard grade $g\in\{A,\ldots,E\}$ off-leaderboard from the fused state; those tables never enter CardScore.
Public ranking uses only $\hat{y}$ under Eqs.~\eqref{eq:score}--\eqref{eq:leadgate}.

\paragraph{Declared FlatVLM contract.}
FlatVLM and ValueAgent share the scorer and raw images, but not intermediate artifacts.
FlatVLM never receives Stage~1 crops, Stage~2 boxes, Expert Knowledge Base dumps, repair-history overrides, or the price table.
Enriched rows that add those artifacts must be labeled explicitly and must not enter the FlatVLM table.
Table~\ref{tab:contracts} records the privilege boundary.
Stages follow warehouse intake: views, defects, screens, fusion, then off-board price.

\begin{table}[t]
\centering\scriptsize
\setlength{\tabcolsep}{3pt}
\renewcommand{\arraystretch}{0.96}
\caption{Declared ranking contracts on PV-Rank.
A check means the artifact may enter the system before CardScore.}
\label{tab:contracts}
\begin{tabularx}{\columnwidth}{@{}Xcc@{}}
\toprule
\textbf{Input artifact} & \textbf{FlatVLM} & \textbf{ValueAgent} \\
\midrule
\rowcolor{qcSand}
Raw order image bag & \cmark & \cmark \\
\rowcolor{qcClay}
Stage~1 canonical crops & \xmark & \cmark \\
\rowcolor{qcClay}
Stage~2 detector boxes or grades & \xmark & \cmark \\
\rowcolor{qcClay}
Stages~4--5 screenshot JSON & \xmark & \cmark \\
\rowcolor{qcClay}
Expert Knowledge Base or repair text & \xmark & \cmark \\
\rowcolor{qcClay}
Numeric price table & \xmark & Stage~6 only \\
\midrule
\rowcolor{qcSlateBest}
Public CardScore on $\hat{y}$ & \cmark & \cmark \\
\bottomrule
\end{tabularx}
\end{table}

\begin{table}[t]
\centering\scriptsize
\setlength{\tabcolsep}{3pt}
\renewcommand{\arraystretch}{1.05}
\caption{ValueAgent inference summary matching Figure~\ref{fig:method}.
The full rule sheet is in the appendix.}
\label{tab:agent-alg}
\begin{tabularx}{\columnwidth}{@{}c l X@{}}
\toprule
\textbf{\#} & \textbf{Stage} & \textbf{Operation} \\
\midrule
\multicolumn{3}{@{}l@{}}{\textit{In:} $X,m$ $\rightarrow$ \textit{Out:} $\hat{y},g,p$} \\
\midrule
\rowcolor{qcSand}
1 & Preprocess & Crop and rotate $X$ into six views $V$ \\
\rowcolor{qcSand}
2 & Object detection & Detector on $V$ yields boxes $B$ and grade \\
\rowcolor{qcTeal}
3 & Scene compliance & Flag or reject if scene is non-compliant \\
\rowcolor{qcTeal}
4 & VLM About & Read About or settings screens into $s_{\mathrm{screen}}$ \\
\rowcolor{qcTeal}
5 & VLM diagnostic & Read diagnostic screens into $s_{\mathrm{diag}}$ \\
\rowcolor{qcSlate}
-- & Fusion + Expert KB & Fuse $B$, grade, screens, KB, and $m$ into $d$ \\
\rowcolor{qcSlate}
-- & MapToScorecard & Map $d$ to scorecard $\hat{y}$ \\
\rowcolor{qcGold}
6 & Price calculation & Off-board price $p$ and grade $g$ from $d$ \\
\bottomrule
\end{tabularx}
\end{table}

\section{Experiments}
\label{sec:exp}

\begin{figure}[t]
\centering
\includegraphics[width=\linewidth]{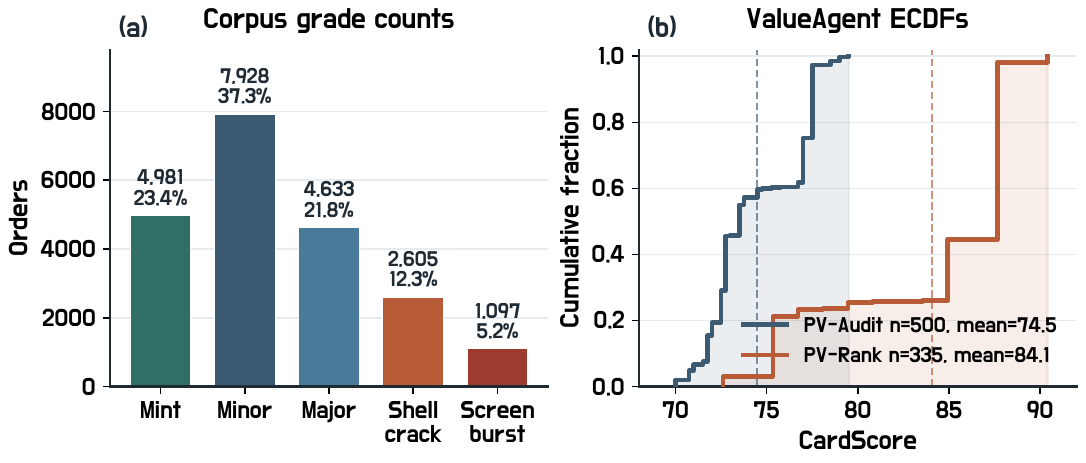}
\caption{Dataset and benchmark statistics on one row.
(a)~PhoneValue-Corpus order counts by cosmetic grade.
(b)~ValueAgent CardScore ECDFs on PV-Audit ($n{=}500$, $\mu{=}74.5$) and PV-Rank ($n{=}335$, $\mu{=}84.1$).
Means are listed in the legend.
PV-Rank is the only public ranking set.}
\label{fig:stats}
\end{figure}

\subsection{Experimental setup}
All leaderboard claims use one PV-Rank cut under LeadGate and one frozen public scorer.
ValueAgent fills the card with the Section~\ref{sec:valueagent} pipeline in Eq.~\eqref{eq:pipeline}. FlatVLM fills the same card from the same raw order images under a single-turn English schema at temperature~0, without detector crops, tools, fusion, or price tables~\cite{qwen25vl2025,gpt4o2024,gemini25pro2025,internvl32025,llavaov2025}.
Table~\ref{tab:flat-contract} states that FlatVLM contract.
The 84.1 versus 69.9 result is a full-system and input-contract comparison, not a controlled test of model architecture or of a single module.
Heuristic optical character recognition and single-image vision-language model runs probe missing modalities under the same scorer.
Neither contract trains on PV-Rank labels: ValueAgent's detector uses PhoneValue-Corpus appearance exports with Rank IDs and image hashes held out, and FlatVLM is evaluated zero-shot on the Rank images.
We report order-level means of CardScore, CosFill, and OpsFill from Eq.~\eqref{eq:fills}.
Bootstrap intervals, severity slices, and off-leaderboard price bands appear below, with aggregation details in the appendix.

Figure~\ref{fig:stats} places Corpus grade mix beside ValueAgent ECDFs.
PV-Audit has mean 74.5 on 500 orders and validates CosGT completeness on a broader mix.
It is not a second leaderboard.
PV-Rank has mean 84.1 on 335 orders and is the only public ranking set.
The Rank ECDF sits higher and tighter, consistent with LeadGate stratification and the appearance-heavy active scorecard with denominator 73.
Minor wear dominates Corpus counts, while PV-Rank deliberately rebalances mint, heavy, and crack or scrap slices so ranking does not collapse to the Corpus mode.

\begin{table}[t]
\centering\scriptsize
\setlength{\tabcolsep}{3.0pt}
\renewcommand{\arraystretch}{0.96}
\caption{FlatVLM evaluation contract on PV-Rank.}
\label{tab:flat-contract}
\begin{tabularx}{\columnwidth}{@{}c l X@{}}
\toprule
\textbf{Step} & \textbf{Component} & \textbf{Setting} \\
\midrule
\rowcolor{qcSand}
1 & Input & all order images and fixed English schema \\
\rowcolor{qcSand}
2 & Decoding & temperature $0$, single turn, no tools \\
\rowcolor{qcTeal}
3 & Parse & structured object to scorecard fields \\
\rowcolor{qcTeal}
4 & Scoring & identical LeadGate CardScore \\
\rowcolor{qcClay}
5 & Exclusions & no detector crops, fusion rules, or price table \\
\bottomrule
\end{tabularx}
\end{table}

\subsection{Main comparison on PV-Rank}
\begin{figure}[t]
\centering
\includegraphics[width=\linewidth]{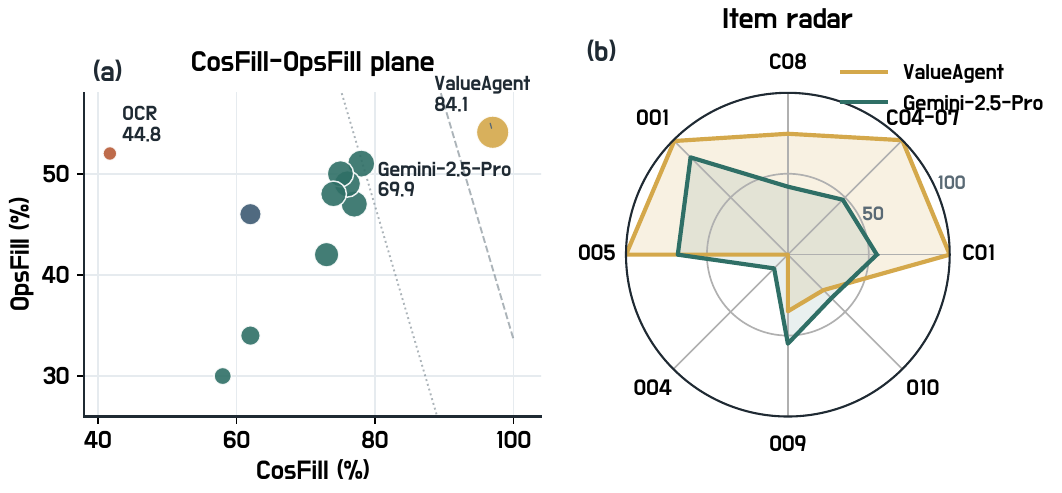}
\caption{Main comparison drawn from Table~\ref{tab:main}.
(a)~CosFill--OpsFill scatter plane with marker size proportional to CardScore and iso-CardScore guides.
(b)~Item-level radar comparing ValueAgent with Gemini-2.5-Pro.}
\label{fig:board}
\end{figure}

ValueAgent reaches CardScore~$=$~\textbf{84.1} with CosFill 97.0 and OpsFill 54.1, as shown in Table~\ref{tab:main} and Figure~\ref{fig:board}.
Operational items trail, especially O04 at 0.0\%, O09 at 35.0\%, and O10 at 31.0\%.
Under Eq.~\eqref{eq:leadgate}, the 14.2-point gap from Gemini-2.5-Pro at 69.9 is mostly cosmetic: CosFill rises from 78.0 to 97.0 (about 13.3 CardScore points), while OpsFill moves only from 51.0 to 54.1 (about 0.9).
Table~\ref{tab:gap} records that arithmetic so the headline gap cannot be misread as verified quote accuracy.
Figure~\ref{fig:board}(a) shows the same split on the CosFill--OpsFill plane, and the item radar in Figure~\ref{fig:board}(b) concentrates ValueAgent's lead on view and regional items while O04, O09, and O10 stay weak for both systems.
Off-leaderboard audits report within-one-tier grade agreement 78\% and price bands 71\% and 84\% at $\pm$10\% and $\pm$20\%, with only moderate CardScore correlation near 0.48.

Among the 14 FlatVLM systems, Gemini-2.5-Pro leads at 69.9, followed by Qwen2.5-VL-72B at 68.0 and GPT-4o at 67.9.
The 14-model mean is 62.4. Model scale alone does not close the observed gap under these declared contracts.
Bag-of-images prompting reaches at most 78\% CosFill while OpsFill can reach about 51\%, which is consistent with screenshot reading without structured cosmetics.
Heuristic optical character recognition at 44.8 and single-image vision-language model at 57.2 show that both modalities matter: OCR misses regional wear, and a single image loses side evidence for C01 and C04--C07.
On the Chinese interface-heavy cut, Doubao scores 67.5, Qwen-72B 68.0, and GPT-4o 67.9; these nearby scores do not establish a language-specific advantage. Qwen's 72B variant exceeds its 7B variant by 14.4 points.
Factor columns in Table~\ref{tab:main} isolate detector crops, screenshot reading, and fusion so ablations and FlatVLM share one comparable schema.
Removing Det lands at 68.8, near the best FlatVLM at 69.9, which is a useful no-detector reference, although metadata and fusion still differ.
Removing VLM or Fuse keeps high CosFill but is associated with lower OpsFill, while Det-only at 74.4 shows crops alone cannot recover staff fields.

\begin{table}[t]
\centering\scriptsize
\setlength{\tabcolsep}{1.4pt}
\renewcommand{\arraystretch}{0.94}
\caption{PhoneValue PV-Rank with 335 orders under one frozen LeadGate scorer.
CardScore$=(51\cdot\mathrm{CosFill}+22\cdot\mathrm{OpsFill})/73$.
Det means Stages~1--2, VLM means Stages~4--5, Fuse means Expert-KB fusion.
This is a full-system and input-contract comparison on the same cut and scorer, not a same-privilege architecture test.}
\label{tab:main}
\begin{tabularx}{\columnwidth}{@{}X*{3}{c}ccc@{}}
\toprule
\textbf{System} & \textbf{Det} & \textbf{VLM} & \textbf{Fuse} & \textbf{Card} & \textbf{Cos} & \textbf{Ops} \\
\midrule
\rowcolor{qcSandBest}
ValueAgent measured & \cmark & \cmark & \cmark & \textbf{84.1} & \textbf{97.0} & \textbf{54.1} \\
\midrule
\rowcolor{qcClay}
w/o Perception-Det & \xmark & \cmark & \cmark & 68.8 & 75.7 & 52.8 \\
\rowcolor{qcClay}
w/o Perception-VLM & \cmark & \xmark & \cmark & 79.3 & 97.0 & 38.2 \\
\rowcolor{qcClay}
w/o Fusion or policy & \cmark & \cmark & \xmark & 80.5 & 94.2 & 48.6 \\
\rowcolor{qcClay}
Det-only, no VLM or fuse & \cmark & \xmark & \xmark & 74.4 & 97.0 & 22.0 \\
\midrule
\rowcolor{qcTeal}
\multicolumn{7}{l}{(A) FlatVLM} \\
\rowcolor{qcTealBest}
Gemini-2.5-Pro & \xmark & \xmark & \xmark & \textbf{69.9} & 78.0 & 51.0 \\
\rowcolor{qcTeal}
Qwen2.5-VL-72B & \xmark & \xmark & \xmark & 68.0 & 77.0 & 47.0 \\
\rowcolor{qcTeal}
GPT-4o & \xmark & \xmark & \xmark & 67.9 & 76.0 & 49.0 \\
\rowcolor{qcTeal}
Doubao-Seed-1.6-Vision & \xmark & \xmark & \xmark & 67.5 & 75.0 & 50.0 \\
\rowcolor{qcTeal}
Claude-Sonnet-4 & \xmark & \xmark & \xmark & 66.2 & 74.0 & 48.0 \\
\rowcolor{qcTeal}
Gemini-2.5-Flash & \xmark & \xmark & \xmark & 63.9 & 72.0 & 45.0 \\
\rowcolor{qcTeal}
InternVL3-78B & \xmark & \xmark & \xmark & 63.7 & 73.0 & 42.0 \\
\rowcolor{qcTeal}
ERNIE-4.5-Turbo-VL & \xmark & \xmark & \xmark & 63.5 & 71.0 & 46.0 \\
\rowcolor{qcTeal}
Hunyuan-Vision & \xmark & \xmark & \xmark & 62.5 & 70.0 & 45.0 \\
\rowcolor{qcTeal}
GLM-4.5V & \xmark & \xmark & \xmark & 60.6 & 69.0 & 41.0 \\
\rowcolor{qcTeal}
Nova-Pro & \xmark & \xmark & \xmark & 59.6 & 68.0 & 40.0 \\
\rowcolor{qcTeal}
Pixtral-Large & \xmark & \xmark & \xmark & 57.6 & 66.0 & 38.0 \\
\rowcolor{qcTeal}
Qwen2.5-VL-7B & \xmark & \xmark & \xmark & 53.6 & 62.0 & 34.0 \\
\rowcolor{qcTeal}
LLaVA-OneVision-72B & \xmark & \xmark & \xmark & 49.6 & 58.0 & 30.0 \\
\rowcolor{qcTeal}
Mean of 14 & \xmark & \xmark & \xmark & 62.4 & -- & -- \\
\midrule
\rowcolor{qcSlate}
Single-image VLM & \xmark & \xmark & \xmark & 57.2 & 62.0 & 46.0 \\
\rowcolor{qcSlate}
Heuristic-OCR & \xmark & \xmark & \xmark & 44.8 & 41.7 & 52.0 \\
\bottomrule
\end{tabularx}
\end{table}

\begin{table}[t]
\centering\scriptsize
\setlength{\tabcolsep}{3.2pt}
\renewcommand{\arraystretch}{0.96}
\caption{LeadGate gap attribution for ValueAgent versus Gemini-2.5-Pro.
$\Delta\mathrm{CardScore}\approx(51\Delta\mathrm{CosFill}+22\Delta\mathrm{OpsFill})/73$.}
\label{tab:gap}
\begin{tabularx}{\columnwidth}{@{}Xccc@{}}
\toprule
\textbf{Term} & \textbf{VA} & \textbf{FlatVLM} & \textbf{$\Delta$ Card} \\
\midrule
\rowcolor{qcSand}
CosFill & 97.0 & 78.0 & $+13.3$ \\
\rowcolor{qcTeal}
OpsFill & 54.1 & 51.0 & $+0.9$ \\
\midrule
\rowcolor{qcSlateBest}
CardScore & 84.1 & 69.9 & $+14.2$ \\
\bottomrule
\end{tabularx}
\end{table}

\subsection{Item-level analysis}
Table~\ref{tab:items} decomposes the scorecard, with item agreement defined as the mean awarded points over the item maximum on orders where the item is active.
ValueAgent is at ceiling on C01 and O05, and strong on O01 at 99.3\%.
FlatVLM lags on C01 at 55.0\% and C08 at 42.0\%, so unordered bags struggle on view and grade-bucket items.
C04--C07 use the public class-and-IoU matcher against CosGT; Rank boxes never train the detector.
On this quality-gated cut, Stage~1 places defects on the correct crop and Stage~2 proposals align with that matcher, and the table reports 100\% for ValueAgent versus 48\% for FlatVLM, which predicts boxes from the raw images.
The 31-point cosmetic block is near saturation for detector-equipped rows, and further CardScore gains must come from OpsFill.
This saturation requires checking coordinate transforms, empty-reference handling, and annotation provenance; the aggregate score alone cannot establish why matching is perfect.
C08 at 74.7\% still reflects cross-taxonomy mapping with adjacent-tier softness.
O04 exact match is 0.0\% for ValueAgent and 12.0\% for FlatVLM, while O05 is 100\% versus 68\%: this pattern is consistent with a taxonomy mismatch, but O05 rewards a one-sided ceiling constraint and does not establish exact grade accuracy or correct visual reasoning.
Under Eq.~\eqref{eq:leadgate}, cosmetics carry about 70\% of LeadGate weight, so the headline gap is cosmetic processing first, and CosFill with OpsFill must stay separate diagnostics.

\begin{table}[t]
\centering\scriptsize
\setlength{\tabcolsep}{2.0pt}
\renewcommand{\arraystretch}{0.94}
\caption{PV-Rank item-level agreement on 335 orders under LeadGate active items.
VA subtotals are $49.47/51$ for C and $11.90/22$ for O.
FlatVLM is Gemini-2.5-Pro.}
\label{tab:items}
\begin{tabularx}{\columnwidth}{@{}Xrrrr@{}}
\toprule
& & \textbf{VA} & \multicolumn{2}{c}{\textbf{Agree (\%)}} \\
\cmidrule(lr){4-5}
\textbf{Item} & \textbf{Max} & \textbf{mean} & \textbf{VA} & \textbf{FlatVLM} \\
\midrule
\rowcolor{qcSand}
C01 view coverage & 7 & 7.00 & 100.0 & 55.0 \\
\rowcolor{qcSand}
C02 class counts & 4 & 3.99 & 99.8 & 61.0 \\
\rowcolor{qcSand}
C03 dark-mode & 3 & 3.00 & 100.0 & 64.0 \\
\rowcolor{qcSandBest}
C04--C07 defect match & 31 & 31.00 & 100.0 & 48.0 \\
\rowcolor{qcSand}
C08 grade agreement & 6 & 4.48 & 74.7 & 42.0 \\
\midrule
\rowcolor{qcTeal}
O01 model & 3 & 2.98 & 99.3 & 85.0 \\
\rowcolor{qcTeal}
O02 capacity & 2 & 1.50 & 75.0 & 70.0 \\
\rowcolor{qcTeal}
O03 IMEI & 1 & 0.80 & 80.0 & 48.0 \\
\rowcolor{qcClay}
O04 cosmetic grade & 6 & 0.00 & 0.0 & 12.0 \\
\rowcolor{qcTealBest}
O05 grade ceiling & 3 & 3.00 & 100.0 & 68.0 \\
\rowcolor{qcTeal}
O06 body flag & 1 & 0.90 & 90.0 & 58.0 \\
\rowcolor{qcTeal}
O07 screen flag & 1 & 0.80 & 80.0 & 52.0 \\
\rowcolor{qcTeal}
O08 water flag & 1 & 0.60 & 60.0 & 40.0 \\
\rowcolor{qcClay}
O09 channel flag & 2 & 0.70 & 35.0 & 55.0 \\
\rowcolor{qcClay}
O10 blockers & 2 & 0.62 & 31.0 & 38.0 \\
\midrule
\rowcolor{qcSlate}
C subtotal & 51 & 49.47 & 97.0 & 78.0 \\
\rowcolor{qcSlate}
O subtotal & 22 & 11.90 & 54.1 & 51.0 \\
\rowcolor{qcSlateBest}
CardScore & -- & -- & \textbf{84.1} & \textbf{69.9} \\
\bottomrule
\end{tabularx}
\end{table}

\subsection{Ablations and stage attribution}
\begin{figure}[t]
\centering
\includegraphics[width=\linewidth]{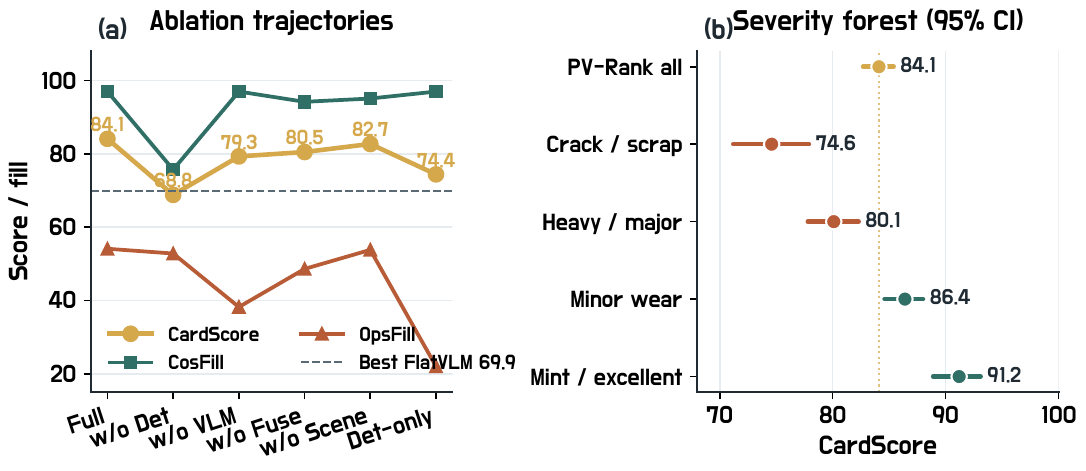}
\caption{Ablations and severity slices.
(a)~CardScore, CosFill, and OpsFill trajectories across ValueAgent ablations, with the best FlatVLM at 69.9.
(b)~Severity forest plot with bootstrap 95\% intervals.}
\label{fig:ablate}
\end{figure}

We next ask which stage is associated with the lift.
Figure~\ref{fig:ablate} and Table~\ref{tab:ablation} report module ablations on the same PV-Rank cut, including the Stage~3 scene-gate variant.
Each row changes one listed factor, with the cut, scorer, and backend fixed, as specified in the appendix.
Removing Stages~1--2 is associated with a drop to 68.8, near the best FlatVLM at 69.9, but that row also removes crops and boxes, so it is not an isolated module proof.
Removing Stages~4--5 yields 79.3 with OpsFill 38.2.
Removing fusion is associated with a 3.6-point drop where screen risk must override an optimistic detector grade.
Detector-only keeps CosFill at 97.0 while OpsFill falls to 22.0.
Removing Stage~3 changes CardScore from 84.1 to 82.7. This comparison alone does not establish that the scene gate improves precision.
Detector-equipped rows keep high CosFill, while screenshot reading and fusion are associated with higher OpsFill.
Because C04--C07 are saturated for detector-equipped variants, the residual open problem is operational-card fidelity under the FlatVLM contract and under ValueAgent's O04, O09, and O10 items.
Single-image evaluation costs 12.7 points versus the full-bag FlatVLM on Gemini-2.5-Pro, echoing the Stage~1--2 ablation.

\begin{table}[t]
\centering\scriptsize
\setlength{\tabcolsep}{3.0pt}
\renewcommand{\arraystretch}{0.96}
\caption{ValueAgent ablation on PV-Rank with 335 orders.
Each row changes one listed factor with the cut and scorer fixed. Drops are associated with that configuration rather than isolated module proofs.}
\label{tab:ablation}
\begin{tabularx}{\columnwidth}{@{}Xccc@{}}
\toprule
\textbf{Variant} & \textbf{CardScore} & \textbf{CosFill} & \textbf{OpsFill} \\
\midrule
\rowcolor{qcSandBest}
ValueAgent full & \textbf{84.1} & \textbf{97.0} & \textbf{54.1} \\
\rowcolor{qcClay}
w/o Perception-Det & 68.8 & 75.7 & 52.8 \\
\rowcolor{qcClay}
w/o Perception-VLM & 79.3 & 97.0 & 38.2 \\
\rowcolor{qcClay}
w/o Fusion or policy & 80.5 & 94.2 & 48.6 \\
\rowcolor{qcClay}
w/o Stage-3 scene gate & 82.7 & 95.1 & 53.8 \\
\rowcolor{qcClay}
Det-only without VLM or fusion & 74.4 & 97.0 & 22.0 \\
\bottomrule
\end{tabularx}
\end{table}

\subsection{Stratified slices and off-leaderboard checks}
Bootstrap 95\% intervals give CardScore $84.1$ in $[82.7,85.4]$ with $B{=}2000$ order-level resamples on PV-Rank only (Figure~\ref{fig:ablate}(b); full slice table in the appendix).
Stratified intervals are computed independently inside each slice, and no interval is reported for PV-Audit or FlatVLM.
Cosmetic strata fall from mint or excellent at 91.2 to crack or scrap at 74.6, and current scores apply to retained Pass orders, not to the raw intake stream.
Off-leaderboard grade within-one-tier is 78\%, and price bands are 71\% and 84\% at $\pm$10\% and $\pm$20\% on the $n{=}298$ non-no-quote subset, against post-intake warehouse quotes (appendix). These checks are not a quoting-accuracy leaderboard.

\subsection{Failure modes and residual gaps}
Three residuals dominate.
O04 stays near zero while O05 is correct, so the card can look cosmetically full while the lexical grade string is still wrong.
O09 and O10 stay weak because channel metadata and blocker vocabulary are sparse.
FlatVLM fails earliest on C01 and C04--C07, and incomplete or Android-heavy intakes remain stress tests outside this Pass board.

\section{Conclusion}
We present PhoneValue-Bench, pairing PhoneValue-Corpus (21{,}550 orders held out from ranking gold) with ValueAgent under declared FlatVLM contracts on one frozen LeadGate scorer.
CardScore with CosFill and OpsFill measures payout-card field-fill completeness, not quote accuracy.
On the compact PV-Rank cut, a full-system and input-contract comparison finds ValueAgent at 84.1 versus 69.9 for the strongest Flat Vision-Language Model, with the largest difference in CosFill, while OpsFill and generalization beyond Pass orders remain limited.
We release the scorer and contracts, while evaluation images and labels remain restricted and the public package alone cannot fully reproduce ranking results.

\section*{LLM Usage Statement}
ChatGPT assisted with manuscript editing, LaTeX formatting, figure typography, and checks of references. The authors take responsibility for the content and results of this work.

{
    \small
    \bibliographystyle{ieeenat_fullname}
    \bibliography{main}
}

\clearpage
\appendix

\begin{center}
{\Large\bfseries Supplementary Material}
\end{center}
\vspace{0.75em}

\section{Overview}
This appendix complements the main paper on the PV-Rank cut of 335 orders.
We first record the dual-ground-truth definition and scorecard item rules, then the shared input-image contract, annotation protocol and remaining agreement gaps, construction filters, isolation scope, fusion priorities, ablation factors, metric aggregation, and model or prompt limits.
CardScore under LeadGate, CosFill, and OpsFill remain the public diagnostics of field-fill completeness.
Grade and price stay secondary metrics outside the public rank.

\section{Broader Impact and Assets}
\paragraph{Broader impact.}
CardScore measures payout-card completeness for remote recommerce intake.
It is not a binding repurchase quote.
ValueAgent at 84.1 and FlatVLM at 69.9 still leave OpsFill near half, so staff review remains necessary for grade strings, channel, and blockers.
Deployments that publish only CardScore without CosFill, OpsFill, and the input contract can overstate readiness.

Photographs may show hands or workplace context.
Consent forms cover the logged orders.

\paragraph{Released assets and data limits.}
The public package includes the frozen scorer, the submission schema and contracts, the fusion priority sheet, aggregate CSV tables, and privacy-checked exterior crops when a safe crop exists.
It does not include the main PV-Rank or PV-Audit images and labels, the full PhoneValue-Corpus, screen captures, serial numbers, account UI, or price slips.
Open FlatVLM checkpoints follow their published licenses and are not redistributed here.
Closed APIs contribute scores only.
PV-Corpus captions and detector sidecars, when access is granted for training, remain under hold-out of Rank IDs and exact image hashes.
The protocol and scorer are inspectable. The main evaluation data remain restricted, so we do not claim that public ranking results can be fully reproduced from the released package alone.

\paragraph{Relation to existing datasets.}
PV-Rank evaluates one intake order as a dual CosGT and OpsGT scorecard.
That differs from single-pane glass inspection in SSGD~\cite{ssgd2023}, controlled anomaly sets such as MVTec AD, VisA, and Kaputt~\cite{mvtec2019,visa2022,kaputt2025}, catalog-attribute resources such as M5Product, SMARTAVE, and ImplicitAVE~\cite{dong2022m5product,wang2022smartave,zou2024implicitave}, and reading or exam suites such as DocVQA, ScreenQA, MMBench, and MMMU~\cite{docvqa2021,screenqa2025,mmbench2024,mmmu2024}.
Those resources are not drop-in substitutes for LeadGate ranking.

\section{Dual Ground Truth and Scorecard Item Rules}
\label{app:scorecard}
Dual ground truth on PhoneValue means two complementary references written on the same intake order, not two equally independent copies of one label set.
CosGT is a vision annotation on the seller photographs: view presence, regional defect boxes, and cosmetic severity, produced by warehouse-trained annotators under a written guideline.
OpsGT is a staff-field annotation on the same order, filled from diagnostic screenshots and intake forms, covering model, capacity, masked International Mobile Equipment Identity last-4, grade strings, risk flags, and blockers.
The two references share one order record and can disagree. They are not interchangeable gold labels.
When they conflict, O-items follow staff priority for O04--O10, and C-items keep vision boxes for C04--C08.
Cosmetic and operational severity conflicts occur on 12.1\% of Rank orders.

Tables~\ref{tab:c-rules}--\ref{tab:o-rules} define each public scorecard item by maximum points, comparison rule, and inactivity condition.
PV-Rank under LeadGate globally deactivates C09, O11--O12, and K01--K02, giving a nominal maximum $y_{\max}{=}73$ before order-specific gates.
Inactive items leave both numerator and denominator.
Inventing missing channels to keep O09 active violates the contract.

\begin{table*}[t]
\centering
\footnotesize
\setlength{\tabcolsep}{3pt}
\renewcommand{\arraystretch}{0.98}
\caption{Cosmetic items C01--C09. IoU threshold $\tau{=}0.5$. LeadGate deactivates C09.}
\label{tab:c-rules}
\begin{tabularx}{\textwidth}{@{}clc>{\raggedright\arraybackslash}X>{\raggedright\arraybackslash}p{0.22\textwidth}@{}}
\toprule
\textbf{ID} & \textbf{Name} & \textbf{Max} & \textbf{Comparison rule vs.\ CosGT} & \textbf{Inactive if} \\
\midrule
\rowcolor{qcSand}
C01 & View coverage & 7 & $7{\times}|V_{\mathrm{pred}}\cap V_{\mathrm{gt}}|/6$ & fewer than 3 mandatory views \\
\rowcolor{qcSand}
C02 & Class counts & 4 & per-class count within $\pm 1$ & no defect classes \\
\rowcolor{qcSand}
C03 & Dark-mode or screen & 3 & exact match on on, off, or dark & no front screen photo \\
\rowcolor{qcTeal}
C04 & Side defect match & 8 & box match with IoU$\ge\tau$ on sides & both side views missing \\
\rowcolor{qcTeal}
C05 & Front defect match & 8 & same on front & no front photo \\
\rowcolor{qcTeal}
C06 & Back defect match & 8 & same on back & no back photo \\
\rowcolor{qcTeal}
C07 & Edge defect match & 7 & same on top or bottom & no edge photo \\
\rowcolor{qcClay}
C08 & Grade agreement & 6 & map worst CosGT to detector bucket, exact=6, adjacent=3 & CosGT grade missing \\
\rowcolor{qcSlate}
C09 & Defect-name recall & 4 & token recall under FullGate only & LeadGate or PV-Rank \\
\bottomrule
\end{tabularx}
\end{table*}

\begin{table*}[t]
\centering
\footnotesize
\setlength{\tabcolsep}{3pt}
\renewcommand{\arraystretch}{0.98}
\caption{Operational items O01--O12 and cross-checks K01--K02. LeadGate deactivates O11--O12 and K.}
\label{tab:o-rules}
\begin{tabularx}{\textwidth}{@{}clc>{\raggedright\arraybackslash}X>{\raggedright\arraybackslash}p{0.22\textwidth}@{}}
\toprule
\textbf{ID} & \textbf{Name} & \textbf{Max} & \textbf{Comparison rule vs.\ OpsGT} & \textbf{Inactive if} \\
\midrule
\rowcolor{qcSand}
O01 & Model identity & 3 & normalized string match & staff model blank \\
\rowcolor{qcSand}
O02 & Capacity & 2 & exact after GB or TB normalize & staff capacity blank \\
\rowcolor{qcSand}
O03 & IMEI or serial & 1 & last-4 after mask & no IMEI in staff or screens \\
\rowcolor{qcClay}
O04 & Cosmetic grade & 6 & exact staff string, strict & staff grade blank \\
\rowcolor{qcTeal}
O05 & Grade ceiling & 3 & predicted ceiling at most staff after A--E map & staff ceiling blank \\
\rowcolor{qcTeal}
O06 & Body flag & 1 & boolean match & flag absent \\
\rowcolor{qcTeal}
O07 & Screen flag & 1 & boolean match & flag absent \\
\rowcolor{qcTeal}
O08 & Water or moisture & 1 & boolean with OR over modalities & not applicable \\
\rowcolor{qcClay}
O09 & Channel or region & 2 & normalized channel match & staff omits channel \\
\rowcolor{qcClay}
O10 & Blockers or no-quote & 2 & set overlap over closed vocabulary & no blocker fields \\
\rowcolor{qcSlate}
O11 & Defect recall ops & 2 & name recall under FullGate & LeadGate or PV-Rank \\
\rowcolor{qcSlate}
O12 & Run sanity & 2 & empty, contradiction, or NaN penalty & LeadGate or PV-Rank \\
\rowcolor{qcGold}
K01 & Vision--ops model & 5 & C-side model vs.\ O01 & LeadGate \\
\rowcolor{qcGold}
K02 & Vision--ops grade & 5 & C08 vs.\ O04 or O05 & LeadGate \\
\bottomrule
\end{tabularx}
\end{table*}

O04 is the strictest lexical item.
Paraphrases that are semantically close still receive zero.
O05 instead scores grade ceilings after the A--E map, so a system may fail O04 yet earn O05 by predicting a sufficiently conservative ceiling. O05 alone does not establish that the economic band is correct.
O03 uses last-4 matching under the masked release.
Full IMEI strings are never published.

\section{Input Image Contract}
\label{app:images}
Both ranking contracts start from the same raw image bag: every image in the order folder after Step~2 anonymization and quality gates, including appearance photographs and diagnostic screenshots.
The reported protocol does not apply a fixed per-order cap such as a three-image budget, and it does not subsample the bag differently for ValueAgent and FlatVLM.
When an order has fewer photographs than the six canonical views, the missing views remain missing. The system does not invent replacement frames, and the corresponding C-items may become inactive under the gates in Table~\ref{tab:c-rules}.
When an order contains more images than are needed to cover the six views plus screenshots, the full remaining bag is still provided to both systems under the reported contract.
Schema examples in the release package include orders with six, seven, or eight images.

Appearance photographs and diagnostic screenshots are not ranked against each other as a selection priority for inclusion in the bag, because both types are already in the folder and both are passed through.
They are used differently after the bag is fixed.
FlatVLM receives the entire bag in one turn and is instructed to use all uploaded images, both cosmetics and screenshots, under the English schema in Table~\ref{tab:direct-prompt}.
ValueAgent may route appearance photographs through Stages~1--3 to produce canonical crops and detector boxes, and may route About, settings, or diagnostic screenshots through Stages~4--5 to produce structured JSON.
Those derived artifacts, together with repair-history text and Expert Knowledge Base records, are privileged inputs that FlatVLM never receives.
A single-image vision-language model probe is a separate modality ablation. It is not the FlatVLM ranking contract.

\section{Annotation Quality and Box Audit}
\label{app:annot}
Eight warehouse-trained annotators label Rank and Audit orders under a written guideline covering view names, defect classes, operational-field definitions, and inactivity conditions.
They first mark which of the six canonical views are present and readable, then draw defect boxes only on present views, then fill operational fields from diagnostic screenshots and intake forms.
When seller free text conflicts with screenshots, screenshots take priority.
Blank fields are preferred to speculative completion.
These labels are reference annotations on the order, not judgments of model outputs.

An approximately 10\% slice is double-labeled independently, with 35 of 335 Rank orders and 50 of 500 Audit orders, before senior adjudication.
On that audit subset, Cohen's $\kappa$ is 0.81 for CosGT severity, 0.88 for view presence, 0.89 for model, 0.94 for capacity, and 0.86 for lock.
The severity $\kappa$ uses the CosGT appearance classes in Table~\ref{tab:taxonomy}.
View presence $\kappa$ uses the six canonical-view present or absent decisions.
Model, capacity, and lock $\kappa$ use the corresponding OpsGT field categories.
About 7.4\% of dual-labeled items require adjudication, mainly boundary severity or missing side views.
Adjudicators may only select published taxonomy buckets.
For boxes, the same IoU$\ge 0.5$ and class-match rule used by C04--C07 decides whether a double-annotated pair agrees.
Pairs that fail that rule enter adjudication and are not averaged into the released CosGT.
A single $\kappa$ is not reported for defect-box matching or for severity-boundary cases beyond the 0.81 severity value, and $\kappa$ confidence intervals are not reported.
PV-Rank severity is unbalanced, with minor wear at 42.4\% and mint or crack or scrap each at 14.3\%, so the severity $\kappa$ should be read with that mix in mind.
Prevalence on the double-label slice itself is not separately tabulated, and a formal statement that annotators were blinded to system outputs is not part of the written protocol beyond the fact that they label reference fields rather than model predictions.

Cosmetic and operational severity conflicts occur on 12.1\% of Rank orders.
O-items follow staff priority for O04--O10, while C-items retain vision GT for C04--C08.

\section{Taxonomy and Construction}
\label{app:taxonomy}
PhoneValue maintains three label spaces: CosGT appearance classes, detector and C08 buckets, and dashboard grades $g\in\{A,\ldots,E\}$, as in Table~\ref{tab:taxonomy}.
C08 scores after this collapse.
O04 ignores the table and requires exact staff strings.
O05 uses the A--E axis.

\begin{table}[t]
\centering
\footnotesize
\setlength{\tabcolsep}{3pt}
\renewcommand{\arraystretch}{0.98}
\caption{Taxonomy crosswalk used by the public scorer.}
\label{tab:taxonomy}
\begin{tabularx}{\columnwidth}{@{}XXX@{}}
\toprule
\textbf{CosGT} & \textbf{Detector or C08} & \textbf{Dashboard $g$} \\
\midrule
\rowcolor{qcSand}
mint & mint & A \\
\rowcolor{qcSand}
minor wear & minor & B \\
\rowcolor{qcTeal}
water ingress & major & C \\
\rowcolor{qcClay}
crack & crack & D \\
\rowcolor{qcClay}
scratch severe & scrap & E \\
\bottomrule
\end{tabularx}
\end{table}

\begin{table}[t]
\centering
\footnotesize
\setlength{\tabcolsep}{3pt}
\renewcommand{\arraystretch}{1.02}
\caption{Construction pipeline for Steps 1--5.}
\label{tab:steps}
\begin{tabularx}{\columnwidth}{@{}c l X@{}}
\toprule
\textbf{Step} & \textbf{Name} & \textbf{Input to output or gate} \\
\midrule
\rowcolor{qcSand}
1 & Collect & raw orders to 21{,}550 image-bearing orders \\
\rowcolor{qcSand}
2 & Preprocess & anonymize, then Pass or Quarantine incomplete evidence \\
\rowcolor{qcTeal}
3 & Label & dual CosGT and OpsGT on Pass orders only \\
\rowcolor{qcTeal}
4 & Corpus & 278{,}748 weak captions and detector sidecars for train \\
\rowcolor{qcClay}
5 & Bench & PV-Audit with 500 orders for integrity, PV-Rank with 335 for ranking \\
\bottomrule
\end{tabularx}
\end{table}

Orders that fail mandatory-view, incomplete staff-text, or unreadable-screenshot gates never receive dual GT and never enter PV-Rank.
PV-Rank is a quality-gated Pass cut, not a random draw from raw intake.
Exact quarantine shares and the pre-anonymization raw-intake distribution before Step~2 are not tabulated in the sources used for this manuscript.
Relative to Corpus, the main paper notes that minor wear dominates Corpus counts, while Rank rebalances mint, heavy, and crack or scrap slices, as listed for Rank in the main composition table.
Diagnostic-screenshot coverage on Rank exceeds 94\% and side-view coverage exceeds 96\%.
Current leaderboard scores apply to these retained Pass orders. They do not automatically extend to quarantined or incomplete intake, or to the raw receiving stream.
Indexed Corpus tiers sum to FullGate 21{,}122, LeadGate 428, and EdgeGate 2 across 21{,}552 rows.
Two non-image folders are excluded from the 21{,}550 image-bearing count.

\section{ValueAgent Fusion Rules}
\label{app:fusion}
Fusion applies deterministic priority rules before \texttt{MapToScorecard}.
The normative sheet used in the reported runs is summarized below.
Full prompt text ships with the release package.

\begin{itemize}
\setlength{\itemsep}{0.15em}
\item \textbf{R1 screenshot identity.} About-screen model and capacity override conflicting view labels.
\item \textbf{R2 IMEI.} Screen last-4 overrides metadata when both exist. Masked release scores last-4 only.
\item \textbf{R3 worst region.} Cosmetic ceiling follows $\mathrm{scrap}\succ\mathrm{crack}\succ\mathrm{major}\succ\mathrm{minor}\succ\mathrm{mint}$.
\item \textbf{R4 water.} Water evidence from diagnostic, repair text, or KB forces ceiling at least major and sets O08.
\item \textbf{R5 scene.} Non-compliant outdoor scenes flag clutter. They do not invent defects.
\item \textbf{R6 channel.} O09 is inactive when staff omits channel. Systems must not invent it.
\item \textbf{R7--R8 flags.} Body and screen flags OR over detector and permitted intake cues; held-out OpsGT must remain inaccessible.
\item \textbf{R9 repair KB.} Repair-text water or foreign-part lexicon can override mint photos.
\item \textbf{R10 blockers.} Activation lock, severe water, foreign parts, or no-quote enter O10 by set overlap.
\item \textbf{R11 MapToScorecard.} Emits $\hat{y}$ without reading the numeric price table.
\item \textbf{R12 price.} Stage~6 price and grade are logged off-leaderboard only.
\end{itemize}

Conflict examples that fire these rules include mint photos versus water text under R4 and R9, capacity 256 versus 128 on screen under R1, and model substring versus About string under R1.
Detector training for the reported runs uses Corpus appearance exports with Rank IDs held out.
C04--C07 on Rank therefore measure alignment with the public matcher, not reuse of Rank boxes at training time.

\section{Isolation Scope}
\label{app:isolation}
The reported isolation claim is an audit on order identifiers and 256-bit image hashes between the training runs used in this paper and the 335-order PV-Rank cut.
Corpus weak labels are never ranking gold.
The detector is trained on PhoneValue-Corpus appearance exports with Rank IDs and exact image hashes held out, and is not fine-tuned on Rank CosGT boxes.
PV-Audit and PV-Rank are treated as disjoint packaged cuts.
This audit does not include a perceptual-hash or near-duplicate search, and it does not quantify crops, recompressed or re-shot frames, or shared product models across train and test.
Exact 256-bit hashes cannot exclude those cases, and shared models are not treated as leakage in the present protocol.

\section{Ablation Protocol}
\label{app:ablation}
Table~\ref{tab:ablation-protocol} names the single factor changed in each ValueAgent variant, with the PV-Rank cut, LeadGate scorer, and reported backend held fixed.
Score changes are associated with that configuration. They do not by themselves prove that one module caused the change in isolation from every other modeling choice, and they are not a same-privilege architecture test against FlatVLM.

\begin{table}[t]
\centering
\footnotesize
\setlength{\tabcolsep}{3pt}
\renewcommand{\arraystretch}{0.98}
\caption{ValueAgent ablation protocol on PV-Rank.
Each row names the factor changed.}
\label{tab:ablation-protocol}
\begin{tabularx}{\columnwidth}{@{}l X@{}}
\toprule
\textbf{Variant} & \textbf{Factor changed} \\
\midrule
\rowcolor{qcClay}
w/o Perception-Det & Skip Stages 1--2 and send raw images to Stages 4--5 \\
\rowcolor{qcClay}
w/o Perception-VLM & Zero About and diagnostic VLM fields \\
\rowcolor{qcClay}
w/o Fusion or policy & Concatenate detector and VLM outputs without R1--R12 \\
\rowcolor{qcClay}
w/o Stage-3 scene gate & Leave the outdoor or street scene gate off \\
\rowcolor{qcClay}
Det-only & Keep Stages 1--2 only, and do not invent staff fields \\
\bottomrule
\end{tabularx}
\end{table}

The w/o Perception-Det row also removes canonical crops and detector boxes, so its drop to 68.8 is associated with that richer change in inputs, not with a frozen image bag under an otherwise identical pipeline.
The w/o Perception-VLM row leaves detector-driven CosFill largely intact and is associated with a lower OpsFill.
The w/o Fusion row preserves raw perception outputs but removes conflict resolution.
Det-only maps Stages~1--2 to cosmetic items and must not copy staff metadata from $m$ into O-items.
A module-contribution claim that isolates architecture from privilege would need the same backbone and the same input artifacts, which these rows do not provide relative to FlatVLM.

\section{Metric Aggregation, Rounding, and Confidence Intervals}
\label{app:metrics}
CardScore, CosFill, and OpsFill are first computed on each order with the gated ratios in the main-paper equations.
Inactive items, including fields with no reference evidence, leave both the numerator and the denominator of that order's ratio.
Blank or unknown predictions do not invent values, and they simply fail to earn the corresponding active points.
The dataset summaries reported in the main tables are order-level means of those per-order ratios, not field-level micro pools that first sum points across all orders and all fields and then divide once.
Item-level agreement, including field accuracy in the item table, is the mean awarded points divided by the item maximum among orders on which that item is active, reported as a percentage.
Group means on severity or family slices use the same order-level mean inside the slice.

Printed table entries are rounded to one decimal for display after that order-level mean is computed.
All reported 95\% intervals are nonparametric percentile intervals for measured ValueAgent CardScore on PV-Rank.
Each of $B{=}2000$ resamples draws orders with replacement and recomputes the same order-level mean, following the construction-appendix bootstrap protocol.
The all-set interval uses sample size 335.
Stratified intervals are computed independently inside each severity or family slice, whose sizes are fixed by the Rank manifest, rather than by post-hoc slicing of the all-set resamples.
The published endpoints match the one-decimal interval table already recorded with the scorer package.
Whether the per-order inputs to that bootstrap were stored at full precision or were already rounded is not separately documented in the manuscript sources, so we do not claim an unrounded-versus-rounded audit beyond those published endpoints.
No confidence interval is reported in the tables for PV-Audit, for FlatVLM, or for CosFill or OpsFill.
PV-Audit remains an integrity cut with mean 74.5 on 500 orders and is not a second leaderboard.

\section{Models, Prompts, and Parse Policy}
\label{app:prompts}
FlatVLM systems are evaluated zero-shot under a frozen contract: the same raw order image bag, temperature $0$, a single turn, no tools, a fixed English schema, and the same LeadGate scorer.
Malformed structured objects zero all active items on that order.
The scorer does not hand-repair free-form prose.
A repair turn is not part of the default FlatVLM contract.
Table~\ref{tab:direct-prompt} abbreviates the English instruction lines used for FlatVLM.
Table~\ref{tab:stage45-prompt} abbreviates the narrower Stages~4--5 screenshot prompt used inside ValueAgent.
The full prompt text is intended to ship with the scorer package and should be pinned by hash.
Exact model snapshot identifiers, API access dates, additional decoding parameters such as token limits, hardware for open-weight runs, and retry counts beyond the malformed-object zeroing rule are not documented in the sources used for this manuscript, and family names alone are not sufficient to re-run the closed APIs bit for bit.

The reported ValueAgent detector is You Only Look Once version~8~\cite{yolov8}, trained on PhoneValue-Corpus appearance exports with PV-Rank order IDs and image hashes held out, and not fine-tuned on CosGT boxes from the ranking cut.
The specific YOLOv8 width variant, weight checksum, and training schedule are not recorded in the manuscript sources.

\begin{table*}[t]
\centering
\footnotesize
\setlength{\tabcolsep}{3pt}
\renewcommand{\arraystretch}{0.98}
\caption{FlatVLM English schema prompt, abbreviated.
The full text is intended to ship with the scorer package.}
\label{tab:direct-prompt}
\begin{tabularx}{\textwidth}{@{}c X@{}}
\toprule
\textbf{\#} & \textbf{Instruction content} \\
\midrule
\rowcolor{qcSand}
1 & Role: score a used-phone trade-in order for a public scorecard. \\
\rowcolor{qcSand}
2 & Input: use all uploaded images for this order, including cosmetics and screenshots. \\
\rowcolor{qcTeal}
3 & Output: return exactly one structured object, with no prose outside fields. \\
\rowcolor{qcTeal}
4 & Required keys: views present, defects with view, class, and box, and a detector grade in mint, minor, major, crack, or scrap. \\
\rowcolor{qcTeal}
5 & Identity keys: model, capacity, IMEI last-4 (do not invent digits), staff grade string, and grade ceiling in \{A,B,C,D,E\}. \\
\rowcolor{qcClay}
6 & Risk keys: flags for body, screen, water, and channel, plus a blockers list from a closed vocabulary. \\
\rowcolor{qcClay}
7 & Unknown policy: leave fields blank or null when unreadable, and never hallucinate IMEI or channel. \\
\rowcolor{qcSlate}
8 & Format policy: structured fields only. Malformed output scores all active items as zero. \\
\bottomrule
\end{tabularx}
\end{table*}

\begin{table*}[t]
\centering
\footnotesize
\setlength{\tabcolsep}{3pt}
\renewcommand{\arraystretch}{0.98}
\caption{ValueAgent Stages 4--5 screenshot-reading prompt, abbreviated.}
\label{tab:stage45-prompt}
\begin{tabularx}{\textwidth}{@{}c X@{}}
\toprule
\textbf{\#} & \textbf{Instruction content} \\
\midrule
\rowcolor{qcSand}
1 & Scope: read only the About, settings, or diagnostic screenshot crop for this turn. \\
\rowcolor{qcTeal}
2 & Extract: model, capacity, IMEI (mask middle digits), battery health, lock status, and channel. \\
\rowcolor{qcTeal}
3 & Output: schema-aligned fields, left blank if the crop is glare, truncated, or non-UI. \\
\rowcolor{qcClay}
4 & Non-goals: do not re-grade cosmetics from this turn. Cosmetic grades come from Stages 1--2. \\
\bottomrule
\end{tabularx}
\end{table*}

\section{Extended Experiments}
\label{app:exp}

\begin{figure}[t]
\centering
\includegraphics[width=\linewidth]{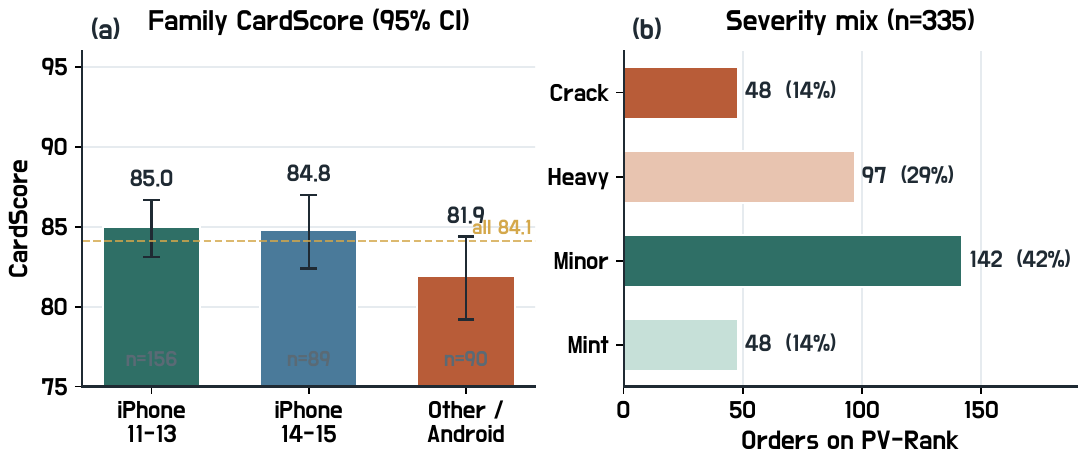}
\caption{Additional PV-Rank slices.
(a)~Device-family CardScore bars with bootstrap 95\% intervals and a dashed all-set mean at 84.1.
(b)~Severity order counts on the 335-order ranking cut.}
\label{fig:app-family}
\end{figure}

\begin{figure}[t]
\centering
\includegraphics[width=\linewidth]{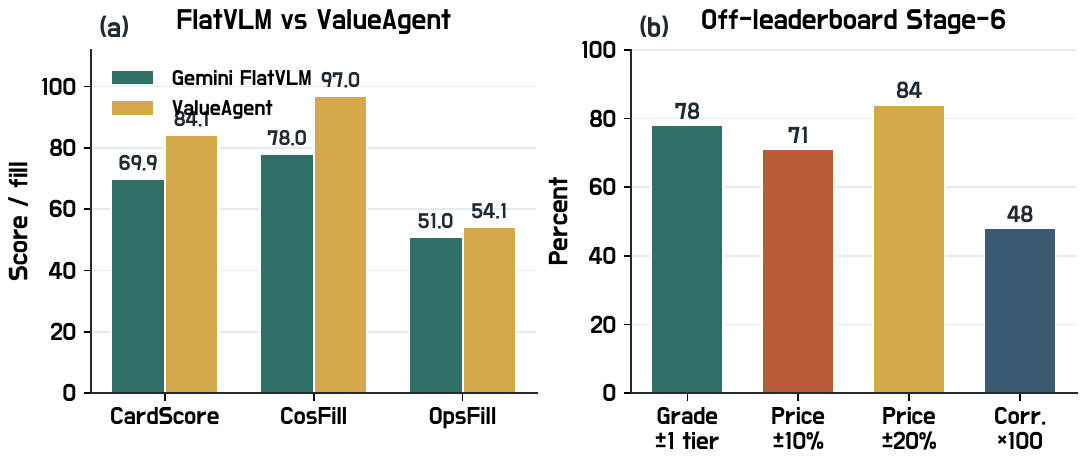}
\caption{Gap attribution and secondary Stage~6 checks.
(a)~CardScore, CosFill, and OpsFill for Gemini FlatVLM versus ValueAgent.
(b)~Off-leaderboard Stage~6 grade and price-band metrics.}
\label{fig:app-gap}
\end{figure}

\paragraph{Matched contracts.}
FlatVLM may consume only the raw order image bag under a single-turn schema at temperature~0.
It must not receive Stage~1 crops, Stage~2 boxes, Expert Knowledge Base dumps, repair-history overrides, or the numeric price table.
ValueAgent may use six-view crops, detector boxes and grades, screenshot JSON from Stages~4--5, and Expert Knowledge Base fusion.
Stage~6 price and dashboard grade remain off the leaderboard for both contracts.
Both still emit the same scorecard $\hat{y}$ for CardScore.

\begin{figure*}[t]
\centering
\includegraphics[width=0.92\textwidth]{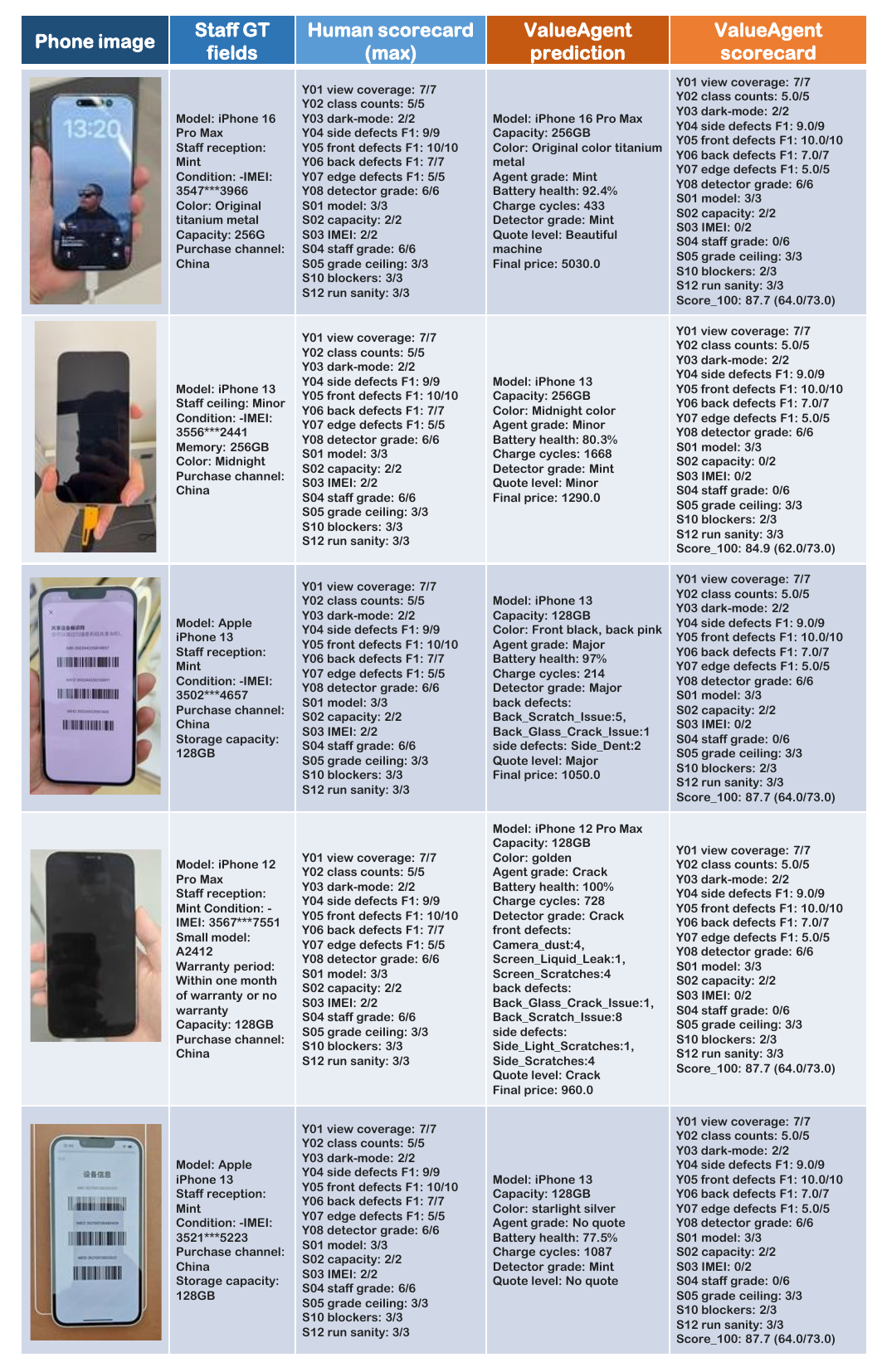}
\caption{Qualitative orders, set 1.
Each row is one order. Columns are the photograph, staff ground-truth fields, the human scorecard, the ValueAgent prediction, and the ValueAgent scorecard.}
\label{fig:appfig1}
\end{figure*}

\begin{figure*}[t]
\centering
\includegraphics[width=0.92\textwidth]{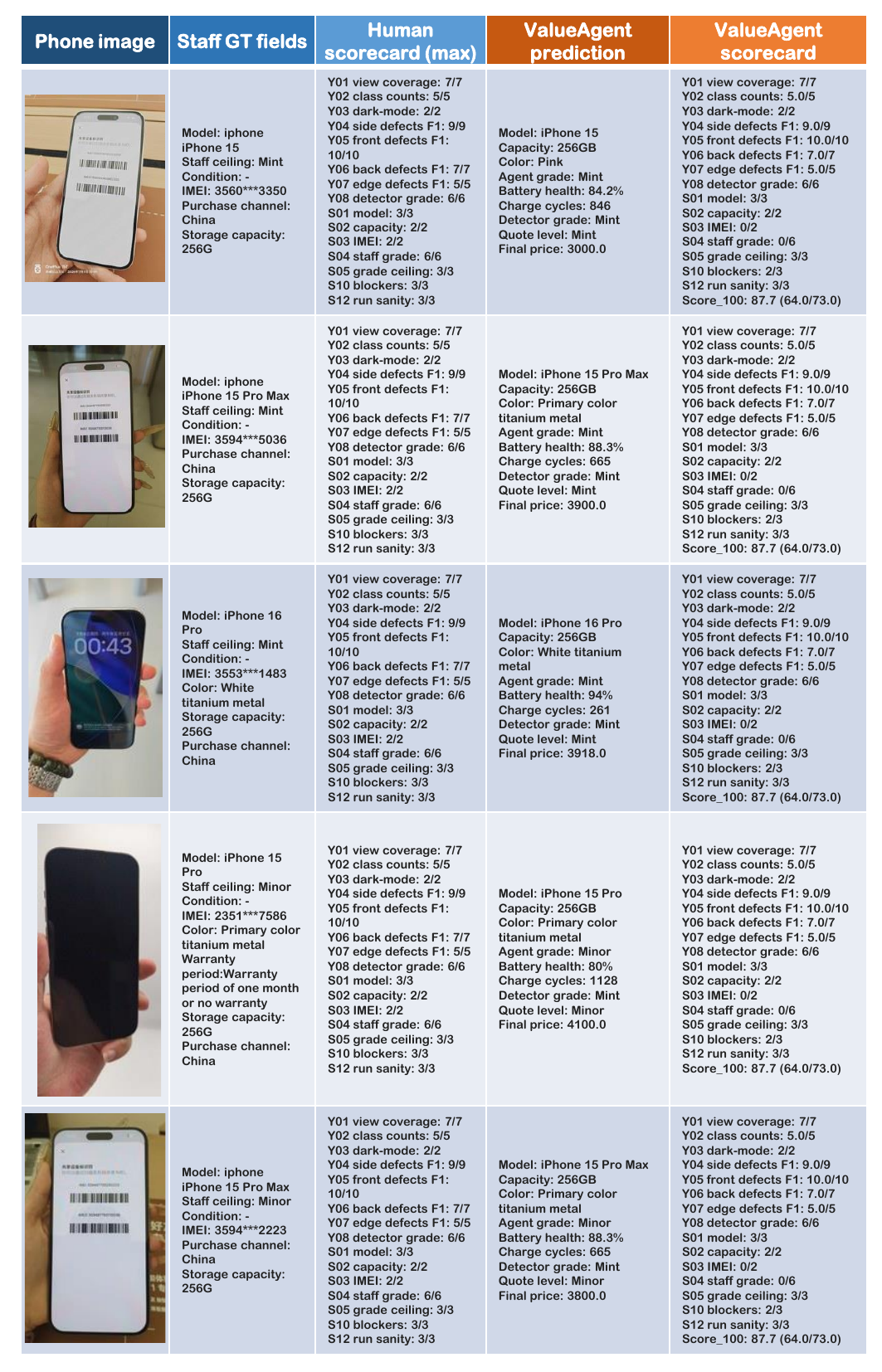}
\caption{Qualitative orders, set 2.
Same five columns as set 1, on a different group of orders.}
\label{fig:appfig2}
\end{figure*}

\paragraph{Order-level case reading.}
Figure~\ref{fig:appfig1} and Figure~\ref{fig:appfig2} place individual orders next to the scorecard that CardScore summarizes.
Each row is one order. The five columns are the exterior photograph, the staff ground-truth fields, the human scorecard, the ValueAgent prediction, and the ValueAgent scorecard.
The two scorecards use the same item order, so an empty cell and a filled cell can be compared without passing through the 335-order mean.
On the printed rows, model, capacity, color, and battery fields are often present in both the staff column and the ValueAgent prediction, while several cosmetic items stay at zero on the human card and on the ValueAgent card together.
A row total near the top of the card is therefore compatible with unrecovered appearance items.
That is the same split the leaderboard reports as OpsFill against CosFill, read here on single orders rather than as an average.
Set~2 keeps the five columns and changes the orders, which shows that the split is not an artifact of one photograph.
The two sets are not a stratified sample, are not bootstrapped, and are not added into the PV-Rank mean.
Hands or a bench edge may appear in the photograph. Identifiers that belong to the logged staff form can appear in the staff column; they are not an extra scored image.

\paragraph{Bootstrap intervals.}
Table~\ref{tab:ci} and Figure~\ref{fig:app-family} report nonparametric bootstrap 95\% intervals for measured ValueAgent CardScore on PV-Rank only, with $B{=}2000$ and sampling with replacement at size 335, as detailed in Section~\ref{app:metrics}.
The all-set mean 84.1 sits in $[82.7,85.4]$.
Cosmetic strata show a severity gradient from mint or excellent 91.2 to crack or scrap 74.6.
Family intervals for iPhone 11--13 and 14--15 overlap.
Other or Android is slightly lower.
Figure~\ref{fig:app-gap} restates the LeadGate gap arithmetic and the off-leaderboard Stage~6 checks.

\begin{table}[t]
\centering
\footnotesize
\setlength{\tabcolsep}{3pt}
\renewcommand{\arraystretch}{0.98}
\caption{Measured ValueAgent CardScore with bootstrap 95\% CIs on 335 orders and $B{=}2000$.}
\label{tab:ci}
\begin{tabularx}{\columnwidth}{@{}Xcc@{}}
\toprule
\textbf{Slice} & \textbf{$n$} & \textbf{Mean and 95\% CI} \\
\midrule
\rowcolor{qcSlateBest}
PV-Rank all & 335 & 84.1 $[82.7, 85.4]$ \\
\rowcolor{qcSand}
Cosmetic mint or excellent & 48 & 91.2 $[88.9, 93.1]$ \\
\rowcolor{qcSand}
Cosmetic minor wear & 142 & 86.4 $[84.6, 88.0]$ \\
\rowcolor{qcClay}
Cosmetic heavy or major & 97 & 80.1 $[77.8, 82.3]$ \\
\rowcolor{qcClay}
Cosmetic crack or scrap & 48 & 74.6 $[71.2, 77.9]$ \\
\rowcolor{qcTeal}
Family iPhone 11--13 & 156 & 85.0 $[83.1, 86.7]$ \\
\rowcolor{qcTeal}
Family iPhone 14--15 & 89 & 84.8 $[82.4, 87.0]$ \\
\rowcolor{qcTeal}
Family other or Android & 90 & 81.9 $[79.2, 84.4]$ \\
\bottomrule
\end{tabularx}
\end{table}

\paragraph{Off-leaderboard grade and price.}
Stage~6 logs coexist with CardScore at 84.1 and correlate only moderately with it.
The card can be mostly filled while the economic band still moves, which is why price is not the public rank and why CardScore is not reported as a quoting accuracy metric.
Within-one-tier grade agreement is computed on PV-Rank orders that have a staff grade.
Price-band hits use relative error against post-intake warehouse quotes, not seller ask prices, at $\pm$10\% and $\pm$20\%.
Orders with O10 no-quote blockers are excluded from the price-band denominator, leaving $n_{\mathrm{price}}{=}298$ on PV-Rank, but they remain in grade audits when a staff grade exists.
The main paper reports the same overall 78\%, 71\%, 84\%, and 0.48 figures.
Slice-wise grade or price percentages that appeared as placeholders in an earlier check report are not repeated here.
Crack and scrap slices are hardest on both grade and price.

\begin{table}[t]
\centering
\footnotesize
\setlength{\tabcolsep}{3.5pt}
\renewcommand{\arraystretch}{0.98}
\caption{Off-leaderboard Stage~6 checks for ValueAgent on PV-Rank.
Grade uses the full cut. Price bands use the $n{=}298$ non-no-quote subset. These metrics are not part of CardScore and are not a quoting-accuracy score.}
\label{tab:biz}
\begin{tabularx}{\columnwidth}{@{}Xc@{}}
\toprule
\textbf{Metric} & \textbf{Value} \\
\midrule
\rowcolor{qcSand}
Grade within one tier & 78\% \\
\rowcolor{qcTeal}
Price within $\pm$10\% & 71\% \\
\rowcolor{qcTeal}
Price within $\pm$20\% & 84\% \\
\rowcolor{qcClay}
Approx.\ CardScore--price-band correlation & 0.48 \\
\bottomrule
\end{tabularx}
\end{table}

\section{Incomplete Inputs and External Validity}
\label{app:robust}
PV-Rank conclusions are limited to quality-gated orders that passed Step~2.
Incomplete uploads with missing views, blurred screenshots, unreadable diagnostics, or conflicting modalities are quarantined from the ranking cut and are not scored as fair peers.
Under a separate robustness probe, systems may abstain or request human review when evidence is insufficient.
An answered CardScore on a damaged upload is not entered on the Rank board.
The board stays compact and comparable. It does not claim coverage of every consumer upload or of the raw receiving stream.
Exact shares of each exclusion reason, and the raw-intake distribution before Step~2, are not tabulated in the sources used for this manuscript.
Cross-device generalization beyond the stratified Rank mix, including harder Android-heavy and incomplete intakes, remains an open stress test.

\section{Scope and Reporting}
CardScore measures scorecard field-fill completeness under LeadGate weights, not end-to-end repurchase pricing or verified quote accuracy.
C04--C07 saturation leaves other active items as possible sources of CardScore gains. Improvements to secondary economic metrics do not change CardScore.
ValueAgent's advantage combines staged perception and fusion under a richer declared contract than FlatVLM.
Among the tested ablations, removing Stages~1--2 has the largest associated CardScore decrease. Input privilege differs between ValueAgent and FlatVLM, so their score gap is a full-system and input-contract comparison, not an architecture-only comparison.
We invite submissions evaluated only on PhoneValue-Bench under the public protocol, not on Corpus weak labels, subject to the data-access limits stated above.

\end{document}